\pdfoutput=1

\documentclass[11pt]{article}

\usepackage[preprint]{neurips_2026}

\usepackage{amsmath,amsfonts,bm}

\def\eqref#1{equation~\ref{#1}}

\def\1{\bm{1}}

\DeclareMathAlphabet{\mathsfit}{\encodingdefault}{\sfdefault}{m}{sl}
\SetMathAlphabet{\mathsfit}{bold}{\encodingdefault}{\sfdefault}{bx}{n}

\usepackage[utf8]{inputenc} 
\usepackage[T1]{fontenc}    
\usepackage{url}            
\usepackage{booktabs}       
\usepackage{longtable} 
\usepackage{amsfonts}       
\usepackage{nicefrac}       
\usepackage{microtype}      
\usepackage{xcolor}         
\usepackage[dvipsnames]{xcolor}
\usepackage{marvosym}
\usepackage{multirow} 
\usepackage{rotating}
\usepackage{graphicx}
\usepackage{wrapfig}
\usepackage{bbm}
\usepackage{subcaption}
\usepackage{algorithm}
\usepackage{algpseudocode}
\usepackage{pgf}
\usepackage{fp}
\usepackage{amssymb}
\usepackage{adjustbox}
\definecolor{cvprblue}{rgb}{0.21,0.49,0.74}
\usepackage{tcolorbox}
\usepackage{listings}
\tcbuselibrary{breakable,listings}
\definecolor{examplebg}{HTML}{F7F7F7}
\definecolor{exampleframe}{HTML}{8A8A8A}
\definecolor{exampletitle}{HTML}{EAEAEA}
\definecolor{figoneblue}{HTML}{0057B8}
\definecolor{figonered}{HTML}{B80000}

\newtcolorbox{paperexample}[1]{
    breakable,
    colback=examplebg,
    colframe=exampleframe,
    colbacktitle=exampleframe,
    coltitle=white,
    fonttitle=\bfseries\normalsize,
    title={#1},
    boxrule=0.6pt,
    arc=1mm,
    left=1.2mm,
    right=1.2mm,
    top=0.8mm,
    bottom=0.8mm,
    before skip=6pt,
    after skip=6pt
}

\lstdefinestyle{paperexamplelisting}{
    basicstyle=\ttfamily\fontsize{6.8pt}{7.8pt}\selectfont,
    breaklines=true,
    breakatwhitespace=false,
    columns=fullflexible,
    keepspaces=true,
    showstringspaces=false,
    escapeinside={(*@}{@*)}
}

\newtcblisting{paperlistingexample}[1]{
    breakable,
    listing only,
    colback=examplebg,
    colframe=exampleframe,
    colbacktitle=exampleframe,
    coltitle=white,
    fonttitle=\bfseries\small,
    title={#1},
    boxrule=0.6pt,
    arc=1mm,
    left=1.2mm,
    right=1.2mm,
    top=0.8mm,
    bottom=0.8mm,
    before skip=6pt,
    after skip=6pt,
    listing options={style=paperexamplelisting}
}

\usepackage{makecell}
\usepackage{enumitem}
\usepackage[colorlinks,citecolor={green!80!black}]{hyperref}
\usepackage{url}
\usepackage{bbm}
\usepackage{subcaption}
\usepackage{algorithm}
\usepackage{algpseudocode}
\definecolor{figoneblue}{HTML}{0057B8}
\definecolor{figonered}{HTML}{B80000}

\title{Mechanistically Interpreting the Role of \\Sample Difficulty in RLVR for LLMs}

\author{Yue Cheng$^{1,2}$\thanks{Equal contribution.} \qquad Jiajun Zhang$^{1*}$\qquad Xiaohui Gao$^{3}$ \\ \textbf{Weiwei Xing$^{1\dagger}$} \qquad \textbf{Zheng Wang$^{4}$} \qquad \textbf{Zhanxing Zhu$^{5}$\thanks{Corresponding authors.}}\\
$^1$Beijing Jiaotong University \quad $^2$AntGroup \quad $^3$Northwestern Polytechnical University \\
$^4$University of Leeds \quad $^5$University of Southampton\\
\texttt{\{yuecheng,jiajunzhang\}@bjtu.edu.cn} \quad \texttt{gaitxh@foxmail.com} \\
\texttt{wwxing@bjtu.edu.cn} \quad \texttt{z.wang5@leeds.ac.uk} \quad \texttt{z.zhu@soton.ac.uk}
}

\begin{document}
\maketitle

\begin{abstract}
  Reinforcement Learning with Verifiable Reward (RLVR) is empirically shown to notably enhance the reasoning performance of large language models (LLMs), particularly in mathematics and programming. However, the mechanistic role of \textit{Sample Difficulty} in RLVR remains poorly understood. In this paper, we investigate RLVR through the lens of difficulty-wise and one-sample analysis. We find that sample difficulty has a non-monotonic effect on RLVR: easy and medium-difficulty problems yield the strongest and most stable reasoning improvements, whereas overly hard problems often provide weak learning signals, induce degenerate behaviors such as answer repetition or skipping necessary computation, and can ultimately degrade the model's pre-existing capabilities. Beyond the obverse of response, we further analyze the model’s internal feature dynamics using Temporal Sparse Autoencoders (T-SAE). Easy problems mainly reinforce direct-answer and basic-computation features while suppressing deliberative-reasoning features; hard problems activate reasoning-related features but become useful only when successful trajectories are sampled; medium-difficulty problems provide a more balanced signal, strengthening both computation and multi-step reasoning features. Motivated by these findings, we propose difficulty-adaptive strategies for hard-sample utilization, using backward-reasoning reformulation and T-SAE-guided training signals to improve reward density and credit assignment during RLVR. Overall, our results identify sample difficulty as a key factor governing both the optimization dynamics and representation evolution of RLVR.
\end{abstract}

\section{Introduction}
Large language models (LLMs) have made significant achievements across a variety of tasks, ranging from mathematics~\cite{liang2025svs,fu2025deepthinkconf}, programming~\cite{luo2025unlocking}, and scientific reasoning~\citep{burgess2026Scientificreason}. Much of this progress has been further amplified by post-training, which adapts pretrained models to elicit stronger reasoning and problem-solving behaviors. Among post-training techniques, Reinforcement Learning with Verifiable Reward (RLVR)~\citep{shao2024deepseekmath,lambert2025rlvr} has emerged as the dominant post-training paradigm, spawning algorithmic innovations such as GRPO~\citep{shao2024deepseekmath}, DAPO~\citep{yu2025dapo}, and VAPO~\citep{yue2025vapo}, as well as extensions to diverse application areas~\citep{xue2025dancegrpo, liu2025flowgrpo, pan2025medvlmr1} and surprising empirical findings~\citep{wang2025reinforcement, zhao2025absolute}. However, most RLVR methods share a fundamental design choice: \textit{they treat samples uniformly regardless of difficulty.} Under group-relative advantage normalization~\citep{shao2024deepseekmath}, only samples that induce mixed rollout outcomes provide meaningful relative-advantage signals. Extremely hard and easy samples therefore contribute little direct learning signal. Although growing empirical evidence suggests that sample difficulty plays an important role in RLVR~\citep{le2026noprompt,tang2026highefficiency,yue2025does,yang2026deepsyn}, the mechanisms by which different difficulty regimes shape optimization dynamics and latent representations remain poorly understood.

Recent work has begun to study this issue from an empirical perspective. Curriculum-based approaches~\citep{luo2025deepscaler} schedule training from easier to harder tasks, online difficulty filters~\citep{bae2026online} prioritize intermediate-difficulty samples, and entropy-guided advantage shaping~\citep{le2026noprompt} seeks to recover useful learning signals from prompts with degenerate reward variance. While these methods lead to practical gains, they primarily operate on outcome-level statistics, such as pass rates, advantage magnitudes, entropy, or reward variance, to select, filter, or reweight samples. As a result, they provide limited insight into how samples of different difficulty levels alter the model's internal computation. This leaves a fundamental question unanswered: \textit{how does sample difficulty shape the model's internal reasoning mechanisms during RL training, and can such a mechanistic understanding inform better training strategies?}

This paper studies how sample difficulty shapes RLVR through one-sample dynamics. Specifically, we begin with controlled subset training and one-sample amplification experiments to characterize how samples of different difficulty levels affect reward dynamics, optimization behavior, and downstream reasoning performance. Our results reveal a non-monotonic effect of difficulty: medium-difficulty samples produce the strongest and most stable gains, whereas very easy and overly hard samples provide weak relative-advantage signals. Hard samples can be especially damaging, as accidentally rewarded trajectories caused by shortcuts or incomplete reasoning may be amplified by group-relative normalization, leading to biased updates that reinforce flawed reasoning behaviors.

Furthermore, this asymmetry is also reflected in the model's internal feature dynamics. 
To probe these dynamics, we introduce a Temporal Sparse Autoencoder (T-SAE) to extract sparse reasoning features from activations along reasoning trajectories. We track the activation frequencies and magnitudes of these features across RLVR training steps, revealing how different difficulty regimes reshape internal reasoning mechanisms.  Easy samples primarily reinforce direct-answer and basic-computation features while suppressing deliberative reasoning; hard samples emphasize complex logical reasoning but weaken computation-related features; medium-difficulty samples strengthen both computation and multi-step reasoning features.  These results suggest that medium-difficulty samples are effective not only because they provide informative reward variance, but also because they induce more balanced reasoning-feature updates.

As a side product of this analysis, we propose two difficulty-aware interventions for improving sample utilization and credit assignment in RLVR. First, we improve the utilization of overly hard samples by reformulating forward problem-solving tasks into backward-reasoning variants, which increases the chance of sampling informative trajectories and usable relative-advantage signals. Second, we introduce Reasoning Feature-Guided Optimization (RFGO), which uses T-SAE reasoning-feature dynamics for mechanistic credit assignment. RFGO reweights token-level updates by future sentence-level feature movement and assigns feature-based proxy rewards only to zero-variance rollout groups. Those interventions support our analysis of hard samples by testing whether the reasoning features they recruit can serve as useful training signals for credit assignment and sample utilization.

\textbf{Related Works.} We discuss related works in Appendix~\ref{relatedwork} due to space limitations.

\section{How Sample Difficulty Shapes Optimization Dynamics?}
\label{sec:impact}
We investigate the role of \textit{sample difficulty} in shaping the optimization dynamics of RLVR. Specifically, we use the empirical success rate of the current policy over $k$ sampled rollouts as a practical proxy for difficulty, and accordingly group samples into easy, medium-difficulty, and hard regimes. To focus on samples that can contribute meaningful policy-gradient updates, we exclude degenerate zero-variance cases in which \textit{all the} rollouts either succeed or fail, as these samples provide no informative relative-advantage signal under group-relative advantage normalization. Section~\ref{sec:group_perspective} studies how different difficulty regimes affect RLVR training at the aggregate level, revealing broad trends in reward evolution, optimization stability, and downstream performance. Section~\ref{sec:per_sample_perspective} then turns to a per-sample analysis, showing that individual samples within the same difficulty regime can still induce highly heterogeneous learning dynamics and failure modes.

\subsection{Preliminaries}
\label{sec:preliminary}
\textbf{Notation.} We denote an LLM policy parameterized by weights $\theta$ as $\pi_\theta$. We sometimes also directly use $\pi_{\mathrm{base}}$ to denote a pre-trained model (e.g., Qwen2.5-math-1.5B, Qwen2.5-math-7B) to contrast its RL counterpart $\pi_{\theta}$. We use $\mathcal{D}$ to denote the query distribution and $\mathcal{V}$ to denote the vocabulary. Given an input query $\boldsymbol{x}$, the model's output is modeled by $\pi_\theta(\cdot|\boldsymbol{x})$. We use $\boldsymbol{y} \sim \pi_\theta(\cdot|\boldsymbol{x})$ to denote the sampling of answer $\boldsymbol{y}$. During RL training, $\pi_{\theta^i}$ denotes the model after the $i$-th policy update.

\textbf{Reinforcement Learning with Verifiable Reward.} 
We formalize the problem of fine-tuning an LLM for complex reasoning tasks under the Reinforcement Learning with Verifiable Reward (RLVR) ~\cite{lambert2025rlvr} paradigm. At each step $t$, the state $s_t=(\boldsymbol{x},\boldsymbol{y}_1,\ldots,\boldsymbol{y}_{t-1})$ is the concatenation of the prompt $\boldsymbol{x}$ and previously generated tokens, and the action $a_t$ is selecting the next token $\boldsymbol{y}_t \in \mathcal{V}$. The LLM with parameters $\theta$ serves as the policy $\pi_\theta$, producing $a_t \sim \pi_\theta(\cdot\mid s_t)$, so that the probability of a trajectory $\boldsymbol{y}=(\boldsymbol{y}_1,\ldots,\boldsymbol{y}_T)$ is
\begin{equation}
    \pi_\theta(\boldsymbol{y} \mid \boldsymbol{x}) = \prod^T_{t=1} \pi_\theta(\boldsymbol{y}_t \mid \boldsymbol{x}, \boldsymbol{y}_{<t}),
\end{equation}
A verifier assigns a sparse, terminal reward $R(\boldsymbol{y})$, which equals $1$ if the final answer extracted from $\boldsymbol{y}$ is correct and $0$ otherwise. Then, the training objective is therefore optimized with policy gradient methods,
\begin{equation}
\mathcal{J}(\theta) = \mathbb{E}_{\boldsymbol{x} \sim \mathcal{D}, \boldsymbol{y} \sim \pi_\theta(\cdot \mid \boldsymbol{x})} \big[ R(\boldsymbol{y}) \big].
\label{eq:rlvrobj}
\end{equation}
The gradient of the objective with respect to the policy parameters $\theta$ can be expressed as
\begin{equation}
    \nabla_\theta \mathcal{J}(\theta) 
    = \mathbb{E}_{(\boldsymbol{x},\boldsymbol{y})\sim\mathcal{D}} \;\mathbb{E}_{\tau\sim\pi_\theta(\cdot|\boldsymbol{x})} \Big[ R(\tau)\nabla_\theta \log \pi_\theta(\tau|\boldsymbol{x}) \Big].
\end{equation}
where gradient coefficient $\mathcal{A}$ depends on the verifier and reward feedback. Detailed policy gradient methods forms, such as GRPO~\cite{shao2024deepseekmath}, are provided in the Appendix~\ref{appendix:core_alg}.

\textbf{Sparse Autoencoders.} Sparse autoencoders (SAEs)~\citep{ng2011sparse,huben2024sparse}  are a class of unsupervised interpretability models that effectively learn sparse feature decompositions of dense LLM activations. Taking in input $\boldsymbol{h}_t \in \mathbb {R}^d$, they map $\boldsymbol{h}_t$ into an overcomplete sparse feature vector 
$\boldsymbol{z}_t\in\mathbb{R}^m$, typically with $m\gg d$, and reconstructs the original activation from this sparse representation:
\begin{equation}
   \boldsymbol{z}_t = f_\theta(\boldsymbol{h}_t) =\sigma(\boldsymbol{W}^{\text{enc}}\boldsymbol{h}_t+\boldsymbol{b}^{\text{enc}}), \qquad \hat{\boldsymbol{h}}_t = \boldsymbol{W}^{\text{dec}}\boldsymbol{z}_t+\boldsymbol{b}^{\text{dec}}
\end{equation}
where $\boldsymbol{W}^{\text{enc}} \in \mathbb{R}^{m \times d}$ is the encoder matrix, and $\boldsymbol{W}^{\text{dec}} \in \mathbb{R}^{d \times m}$ is the decoder matrix. The encoder and decoder bias are $\boldsymbol{b}^{\text{enc}} \in \mathbb
{R}^m$ and $\boldsymbol{b}^{\text{dec}} \in \mathbb{R}^d$, respectively. Temporal Sparse Autoencoders (T-SAEs)~\cite{bhalla2026tsae} extend standard SAEs by explicitly leveraging the sequential structure of language. With the assumption that the first $l$ indices are relatively stable high-level features $\boldsymbol{z}^{0:h}_t$ and the last $m-l$ indices are rapid change low-level features $\boldsymbol{z}^{h:m}_t$, the reconstruction component of the T-SAEs objective can therefore be written as
\begin{equation}
    \mathcal{L} = \| h_t - \boldsymbol{W}^{\text{dec}}_{0:h}\boldsymbol{z}^{0:h}_t - \boldsymbol{b}^{\text{dec}}\|_2^2 + \| h_t-\boldsymbol{W}^{\text{dec}}z_t - \boldsymbol{b}^{\text{dec}}\|.
\end{equation}
See more details about T-SAEs in the Appendix~\ref{appendix:core_alg}.

\subsection{How Difficulty Regimes Shape Optimization Dynamics}
\label{sec:group_perspective}

For each training sample $\boldsymbol{x}$, we sample $k$ responses (rollouts) $\{\boldsymbol{y}_i\}_{i=1}^{k}$ from the current policy and assign each rollout a binary verification reward $R_i \in \{0,1\}$. We use the per-sample empirical success rate $\texttt{success}@ k  = \frac{1}{k}\sum_{i=1}^{k} R_i$ as a practical proxy for sample difficulty. Samples with high success rates $\texttt{success}@ k = 1$ are categorized as $\texttt{easy}@ k$ regimes, those with intermediate success rates $0 \leq \texttt{success}@ k \leq 1$ as $\texttt{medium}@ k$, and those with low success rates $\texttt{success}@ k = 0$ as $\texttt{hard}@ k$. See Appendix~\ref{app:task_difficulty} for more details about the definition of sample difficulty.

In the context of GRPO~\cite{shao2024deepseekmath}, rollout advantages are normalized within each group $\hat{A}^i_t = \nicefrac{R^i - \mathrm{mean}(\{R^i\}_{i=1}^k)}{\mathrm{std}(\{R^i\}_{i=1}^k)}$. When all rollouts receive the same reward, the within-group reward variance becomes zero, and the sample provides no informative relative-advantage signal. Following~\citep{le2026noprompt}, we refer to these cases as \textit{zero-variance samples}. 
We exclude strictly zero-variance samples from the following case studies, since they do not induce non-trivial reward-driven updates. We empirically validate this choice in Appendix~\ref{app:zero_variance}. Exploiting 512 rollouts from Qwen2.5-Math-1.5B, we extract a zero-variance subset from MATH. We point out that their advantage signals vanish, leaving the policy update dominated by KL regularization rather than reward-driven reinforcement. Having removed these degenerate cases, we now turn to the central question of this section: \textit{how do non-zero-variance samples of different difficulty levels affect RLVR training? }

\begin{figure}[htbp]
    \vspace{-15pt}
    \centering
    \includegraphics[width=0.90\linewidth]{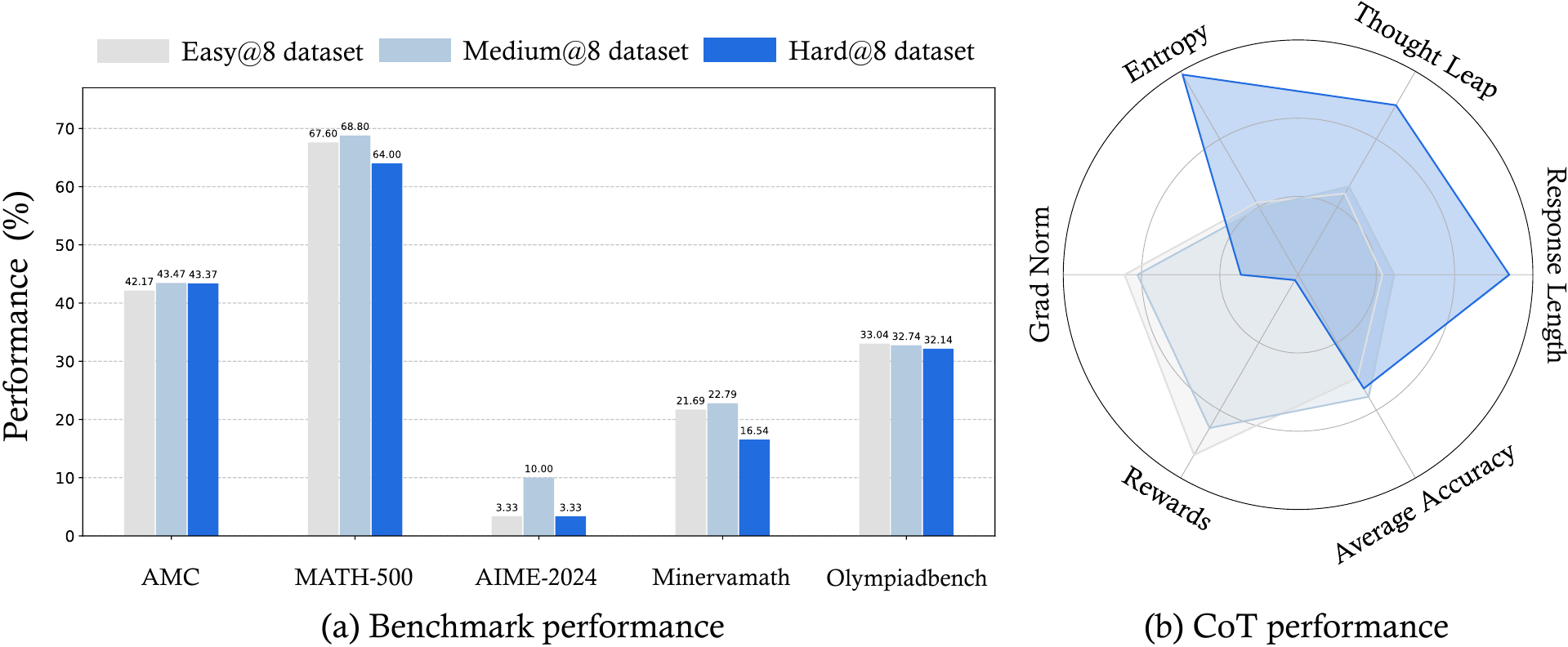}
    \caption{Overall performance of GRPO across multiple benchmarks using Qwen2.5-Math-1.5B as the base model. We partition training samples into \textbf{Easy}, \textbf{Medium}, and \textbf{Hard} regimes, where Easy is defined as \texttt{easy}@8, Hard as \texttt{hard}@8, and Medium contains the remaining samples.}
    \label{fig:subset_grpo}
    \vspace{-0pt}
\end{figure}

\textbf{Empirical Observations on Sample Difficulty.} As a first step, we use \texttt{success}@8 as a coarse proxy for sample difficulty, matching the group size used in GRPO.  We split the training data into \texttt{easy}@8, \texttt{medium}@8, and \texttt{hard}@8 regimes, and train Qwen2.5-Math-1.5B and Qwen2.5-Math-7B on each subset to isolate the effect of difficulty, as shown in Figure~\ref{fig:subset_grpo}. Across both base models, we find a clear non-monotonic trend: Medium samples yield the largest gains, Easy samples provide smaller but stable improvements, and Hard samples are the least effective. The strongest evidence comes from mixed-difficulty training: "\texttt{Easy+Medium}" consistently achieves the best performance and even outperforms the full dataset. This indicates that overly hard samples can dilute the useful learning signal in RLVR, motivating a more fine-grained analysis of sample difficulty beyond the coarse partition. More comprehensive comparisons are provided in Appendix~\ref{app:diff_groups}.

\begin{figure}[bp]
    \vspace{-10pt}
    \centering
    \includegraphics[width=1.0\linewidth]{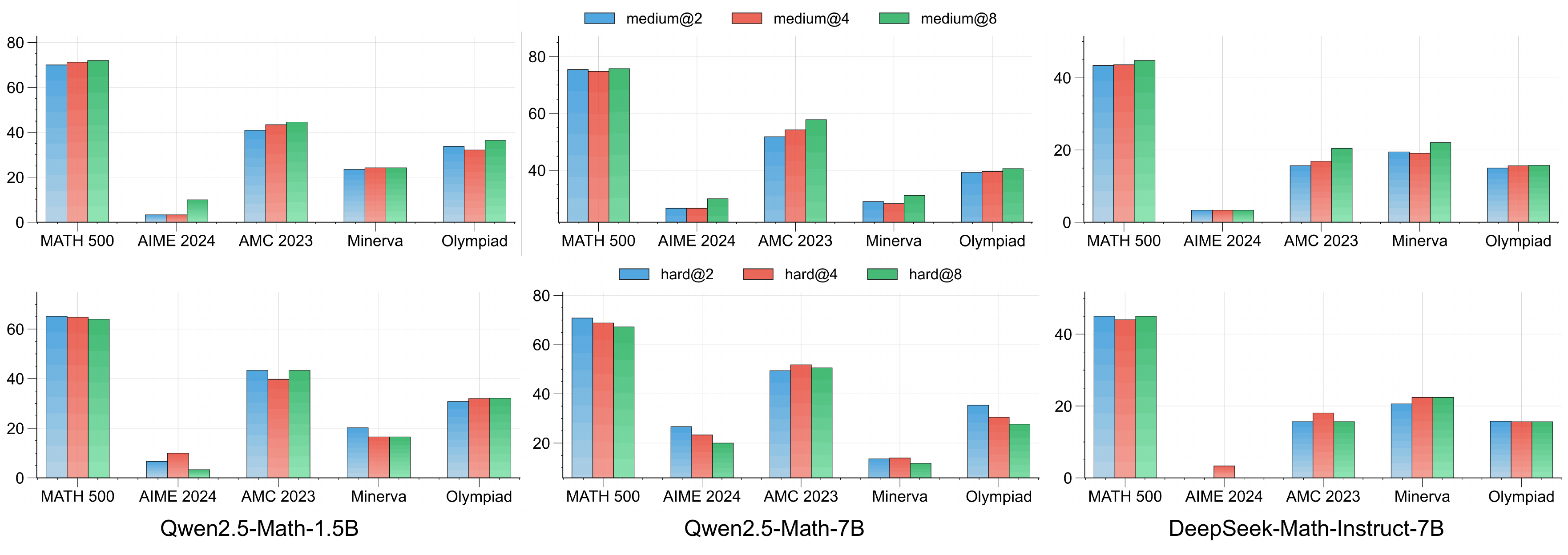}
    \caption{Difficulty curriculum for practical RLVR. We compare training on curriculum subsets defined by different rollout-based difficulty thresholds across three base models. Samples that are solvable within a limited rollout budget consistently yield stronger improvements than persistently unsolved samples, highlighting the importance of difficulty-aware data selection in RLVR.}
    \label{fig:diff_curr}
    \vspace{-15pt}
\end{figure}

\textbf{Difficulty Curriculum for Practical RL.} 
The coarse split suggests that overly difficult samples can dilute RLVR training, but it does not reveal how the training signal changes across finer difficulty regimes. We therefore construct a finer difficulty curriculum by varying the rollout budget used to determine whether a sample is solvable by the base policy. For each sample $\boldsymbol{x}$, we sample $k=8$ rollouts from $\pi_{\mathrm{base}}$ and obtain binary rewards $R_i(\boldsymbol{x}) \in \{0,1\}$. 
For $k \in \{2,4,8\}$, we define
\begin{equation}
    \texttt{medium}@k
    =
    \{
    \boldsymbol{x} \in \mathcal{D}:
    \sum_{i=1}^{k} R_i(\boldsymbol{x}) \geq 1
    \},
    \qquad
    \texttt{hard}@k
    =
    \{
    \boldsymbol{x} \in \mathcal{D}:
    \sum_{i=1}^{k} R_i(\boldsymbol{x}) = 0
    \}
\end{equation}
Here, $\texttt{medium}@k$ contains samples solved at least once within $k$ attempts, while $\texttt{hard}@k$ contains samples that remain unsolved within the same budget. As $k$ increases, $\texttt{medium}@k$ progressively includes harder but still solvable samples, $\texttt{medium}@2 \subseteq \texttt{medium}@4 \subseteq \texttt{medium}@8$. Conversely, $\texttt{hard}@k$ progressively removes relatively easier failures and retains more persistently unsolved samples,
$\texttt{hard}@8 \subseteq \texttt{hard}@4 \subseteq \texttt{hard}@2$. Since rollout sampling is stochastic, $\texttt{hard}@8$ only indicates a low probability of receiving a positive verifier reward under the reference policy and rollout budget.

We train each base model on the resulting subsets and report results in Figure~\ref{fig:diff_curr}. 
On the solvable side, increasing $k$ enlarges $\texttt{medium}@k$ by adding harder but still occasionally solvable samples, and performance almost always improves monotonically. This indicates that RLVR benefits from medium-difficulty samples as long as the base model can still solve them occasionally. The hard side shows a different pattern. Although increasing $k$ makes $\texttt{hard}@k$ smaller and more persistently unsolved, performance does not degrade monotonically. Instead, results are unstable across models and benchmarks, suggesting that hard samples are heterogeneous: they may either trigger useful policy updates or contribute noisy, uninformative gradients. Detailed per-benchmark numbers are provided in Table~\ref{tab:one_sample_individual_benchmarks}. These findings show that aggregate difficulty is insufficient to explain the effect of hard data, motivating a sample-level analysis of \textit{how individual samples steer reasoning behavior during RLVR.}

\begin{figure}[t]
    \vspace{-15pt}
    \centering
    \includegraphics[width=\linewidth]{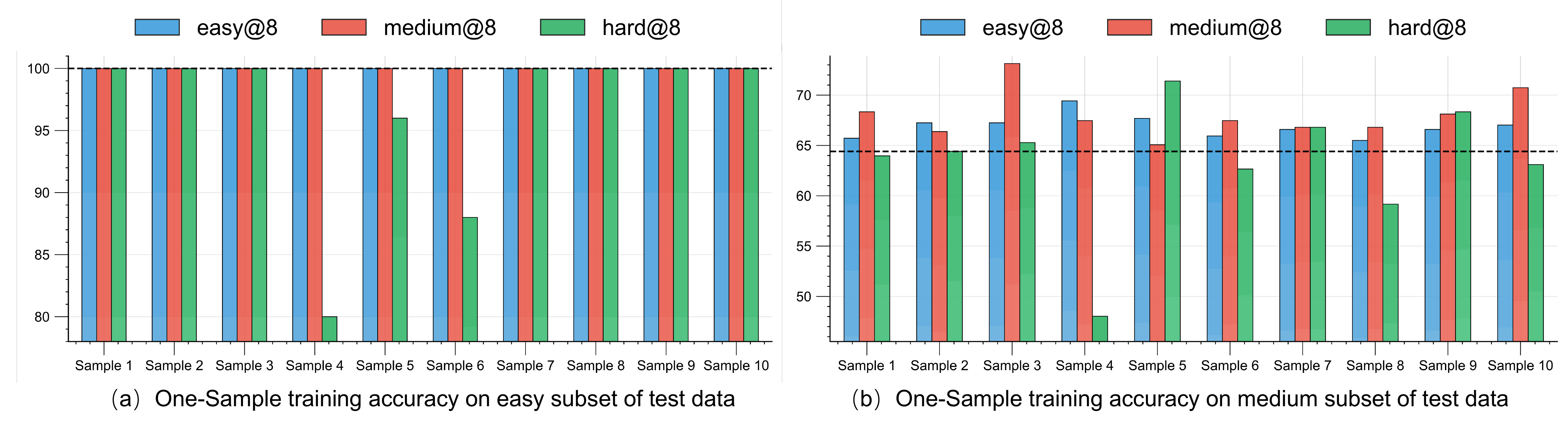}
    \caption{One-sample RL performance on MATH-500 test subsets.  For each training regime (\texttt{easy}@8, \texttt{medium}@8, and \texttt{hard}@8), we sample 10 individual problems, train on each problem separately, and evaluate on the MATH-500 \texttt{easy}@8 and \texttt{medium}@8 test subsets. The \texttt{hard}@8 test subset is omitted because all runs achieve $0$ accuracy.}
    \label{fig:one_sample}
    \vspace{-15pt}
\end{figure}

\subsection{How Individual Samples Shape Optimization Dynamics}
\label{sec:per_sample_perspective}
The curriculum-level analysis shows that hard data behaves irregularly in aggregate. We next ask whether this irregularity already appears at the level of a single training sample. Even within the same difficulty regime, different samples may reward different trajectories and induce very different policy updates. We therefore move from curriculum-level comparisons to one-sample RL dynamics.

\textbf{One-Sample Amplification.} To isolate the effect of an individual sample, we train the policy on one sample at a time while keeping the standard GRPO group-rollout structure. For a selected sample $\boldsymbol{x}$, each update samples $G$ rollouts from $\pi_{\theta^t}$ and computes group-relative advantages from verifier rewards. This removes aggregation across samples while preserving within-sample credit assignment. At step $t$, we track the sample reward $\bar{R}_t(\boldsymbol{x}) = \frac{1}{G}\sum_{i=1}^{G} R(\boldsymbol{y}_i^t)$, where $\qquad \boldsymbol{y}_i^t \sim \pi_{\theta^t}(\cdot \mid \boldsymbol{x})$, the KL divergence from the reference policy $\mathcal{D}_{\mathrm{KL}}^t(\boldsymbol{x}) = \mathbb{E}_{\boldsymbol{y}\sim \pi_{\theta^t}(\cdot \mid \boldsymbol{x})} \left[ \log \frac{ \pi_{\theta^t}(\boldsymbol{y}\mid \boldsymbol{x}) }{\pi_{\mathrm{ref}}(\boldsymbol{y}\mid \boldsymbol{x}) } \right]$ and downstream \texttt{pass}@1 accuracy. Together, these metrics measure whether the sample becomes solvable, how strongly it moves the policy, and whether the induced update transfers beyond the training sample.

\begin{wrapfigure}{r}{0.5\textwidth}
    \vspace{-10pt}
    \centering
    \includegraphics[width=1\linewidth]{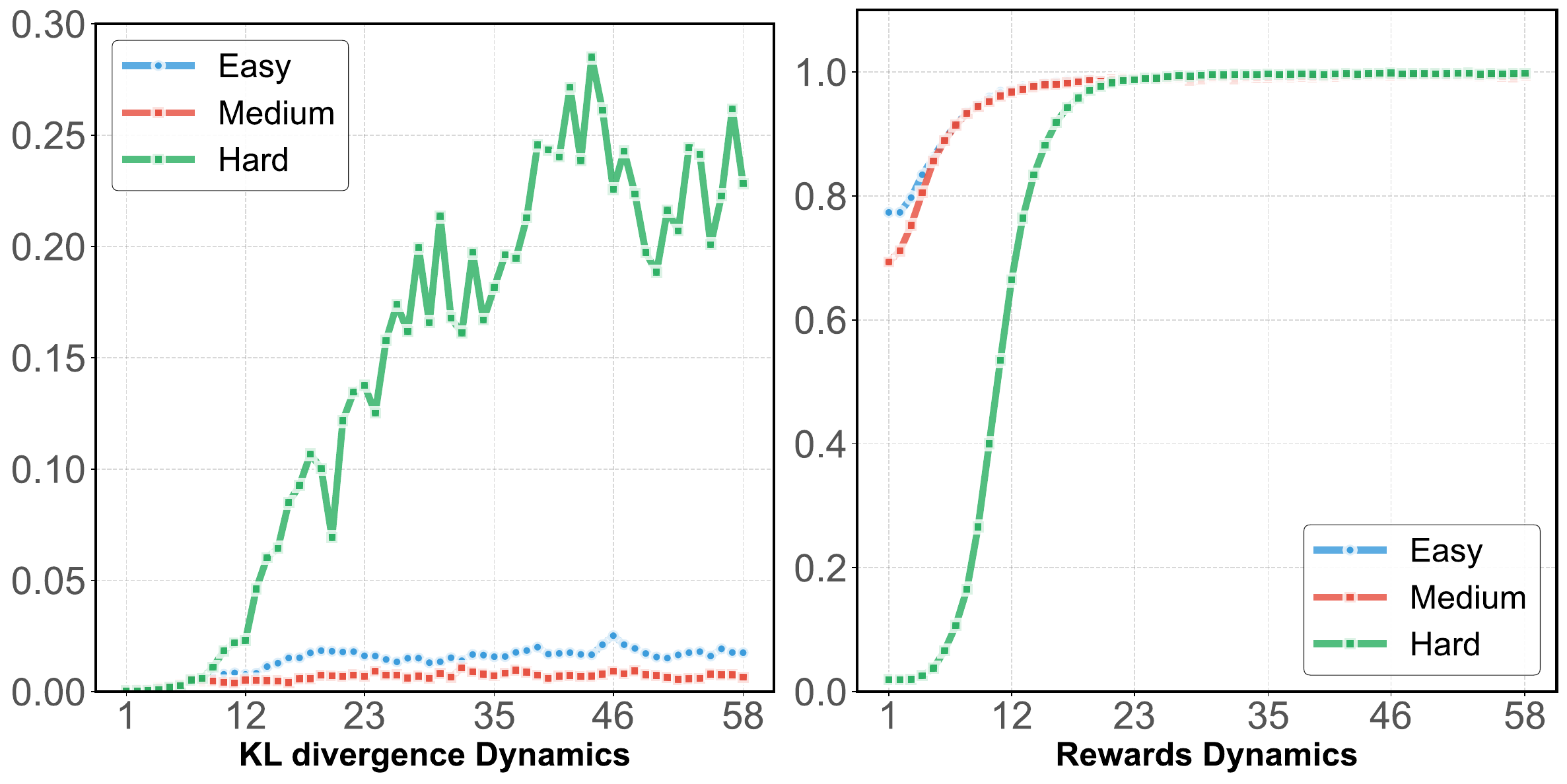}
    \caption{Representative reward and KL dynamics for one-sample RL on three examples selected from \texttt{Easy}@8, \texttt{Medium}@8, and \texttt{Hard}@8.}
    \label{fig:one_dynamics}
    \vspace{-15pt}
\end{wrapfigure} 

\textbf{Difficulty-Dependent Optimization Dynamics.} Figure~\ref{fig:one_sample} shows that easy and medium samples induce relatively stable updates. Training on \texttt{Easy}@8 or \texttt{Medium}@8 usually preserves or improves downstream accuracy, suggesting that RL reinforces useful behavior when the base model can already solve the sample with non-negligible probability. Hard samples are much less predictable. Although some \texttt{Hard}@8 samples improve downstream performance, many sharply degrade it, sometimes below the base model. Their effect therefore depends not only on difficulty, but on which rare rewarded trajectories are amplified by GRPO. Figure~\ref{fig:one_dynamics} illustrates this mechanism: the hard sample reaches high training reward only with much larger KL growth, whereas easy and medium samples improve with more controlled policy movement. Thus, high reward on a hard sample need not indicate useful learning; it may instead reflect accidental correctness or shortcut reasoning.

Overall, \texttt{Hard}@8 obtains a lower average score of $32.61$ and much larger variance, spanning both strong improvement and severe degradation. Appendix~\ref{appendix:one_sample_dynamics} further tracks downstream test dynamics and shows that harmful hard samples often cause collapse only after their effects accumulate over training. This asymmetry explains why hard data is unstable: reward improvement on easy and medium samples tends to transfer, while reward improvement on hard samples can amplify brittle behaviors and damage existing capabilities.

\textbf{Failure Modes of Harmful Hard Samples.} We further inspect harmful hard samples and find that their rewards often come from spurious solution patterns rather than valid reasoning. Example~\hyperref[ex:mem]{2.1} illustrates an answer-only shortcut: the model directly outputs “$\texttt{\textbackslash boxed\{40\}}$” for a geometry problem that should require using $l\|k$, the labeled $130^\circ$ angle, and the right-angle constraint. Outcome-only verification still rewards this rollout because the final answer is correct, despite the absence of supporting reasoning. Appendix~\ref{appendix:failure_examples} shows that this issue is broader: harmful hard samples can involve goal misunderstanding, question--answer mismatch, missing aggregation, calculation or execution errors, inconsistent output formats, and logical constraint errors. Thus, RL on such samples may amplify brittle shortcuts or artifacts rather than transferable reasoning.

\phantomsection\label{ex:mem}
\begin{paperlistingexample}{Example 2.1: \textit{Spurious Reward from Answer-Only Output}}
(*@\textnormal{\bfseries Question:}@*) In the diagram, $l\|k$. What is the number of degrees in $\angle SRQ$? [asy] draw((-.4,-.4)--(2,2)--(2,-.4)); draw((-.5,0)--(3,0),Arrows); draw((-.5,1)--(3,1),Arrows); draw((1.9,0)--(1.9,.1)--(2,.1)); label("$S$",(1,1),NNW); label("$R$",(2,2),N); label("$Q$",(2,1),NE); label("$l$",(3,1),E); label("$k$",(3,0),E); label("$130^{\circ}$",(1,1),SSE); [/asy] 
(*@\textnormal{\bfseries\color{figonered} Answer:}@*) (*@\textcolor{figonered}{$40^\circ$}@*)
(*@\textnormal{\bfseries\color{figoneblue} Response:}@*)  (*@\textcolor{figoneblue}{The number of degrees in \texttt{\textbackslash angle SRQ} is \texttt{\textbackslash boxed\{40\}}.}@*)
\end{paperlistingexample}

This explains why reward-flipped hard samples can be especially risky. All-zero hard samples induce little reward-driven learning, whereas reward-flipped hard samples can produce large updates based on accidental correctness, shortcut reasoning, or verifier artifacts. Therefore, one-sample RL reveals the fine-grained source of hard-data instability: hard samples are heterogeneous, and many of them disrupt rather than improve the model's existing reasoning ability. This motivates our next feature-level analysis, where we use T-SAE to examine how different samples reinforce or suppress internal reasoning features during RLVR.

\section{How Sample Difficulty Reshapes Reasoning Features}
\label{sec:tsae_analysis}
Sample difficulty induces distinct optimization dynamics during RLVR. However, these optimization signatures do not directly explain how difficulty reshapes the model’s internal reasoning mechanisms. To bridge this gap, we introduce a feature-dynamics framework based on a Temporal Sparse Autoencoder (T-SAE) probe in Section~\ref{sec:sae_analysis}. Concretely, we apply the same T-SAE model \textit{across different training steps} to obtain a shared sparse feature basis, and we then track how reasoning-relevant features change over time. This setup enables us to quantify whether different difficulty regimes reinforce, suppress, or reorganize internal reasoning features during training. We provide the concrete experimental design and the corresponding comparison results in Section~\ref{sec:experiment}.

\subsection{Tracking Reasoning Features with a T-SAE Probe}
\label{sec:sae_analysis}
In this section, we track how sparse features evolve along the RLVR training trajectory. Since standard SAEs treat token activations independently, they can miss the sequential structure of reasoning traces. T-SAE addresses this limitation by incorporating temporal consistency and sparse top-$k$ coding, yielding features better aligned with temporally extended reasoning behavior. We therefore use a frozen T-SAE as a shared probe for analyzing reasoning-feature dynamics across training steps. Concretely, for a policy $\pi_{\theta^\tau}$ at training step $\tau$, let $\boldsymbol{h}_{t,\tau}^{\ell} \in \mathbb{R}^{d}$ denote the hidden activation at token position $t$ from layer $\ell$. The T-SAE maps this activation to a sparse latent code:
\begin{equation}
    \boldsymbol{z}_{t,\tau} = \zeta(\boldsymbol{h}_{t,\tau}), \qquad \hat{\boldsymbol{h}}_{t,\tau} = \xi(\boldsymbol{z}^{0:h}_{t,\tau})
\end{equation}
where $\zeta(\cdot)$ and $\xi(\cdot)$ denote the encoder and decoder, respectively. Each coordinate of $\boldsymbol{z}_{t,\tau} \in \mathbb{R}^{m}$ corresponds to a sparse latent feature, and its value indicates the activation strength of that feature at token position $t$. This representation allows us to compare reasoning-related activation patterns in a common sparse basis rather than in the original dense hidden space.

Given a sequence of policies $\{\pi_{\theta^\tau}\}$ produced during RLVR, our goal is to characterize how the corresponding sparse features evolve as $\tau$ changes. We then adopt \textsc{ReasonScore}~\citep{galichin2026have} to quantify whether a feature is associated with reasoning behavior.

Let $\mathcal{V}_{\mathrm{reason}}=\{v_r\}_{r=1}^{R}$ denote a predefined collection of reasoning-token groups. For feature $i$, regime $d$, and training step $\tau$, let $\mu^{d,+}_{i}(\tau)$ be the mean activation of feature $i$ on reasoning tokens, let $\mu^{d,-}_{i}(\tau)$ be the mean activation on non-reasoning tokens, and let $H^{d}_{i}(\tau)\in[0,1]$ be the normalized entropy of feature $i$ across the $R$ reasoning-token groups. We then define
\begin{equation}
    \textsc{ReasonScore}^{d}_{i}(\tau)
    =
    \frac{\mu^{d,+}_{i}(\tau)}{\sum_{j}\mu^{d,+}_{j}(\tau)}
    \cdot
    \big(H^{d}_{i}(\tau)\big)^{\alpha}
    -
    \frac{\mu^{d,-}_{i}(\tau)}{\sum_{j}\mu^{d,-}_{j}(\tau)},
    \label{eq:reasonscore}
\end{equation}
where $\alpha$ controls the strength of the entropy normalization. Intuitively, a feature receives a high score when it activates strongly and broadly across reasoning tokens while remaining selective against non-reasoning contexts. For each feature $i$ and regime $d$, we then track the trajectory $\{\textsc{ReasonScore}^{d}_{i}(\tau)\}_{\tau \in \mathcal{T}_{d}}$ over the available checkpoints $\mathcal{T}_{d}$. Comparing these trajectories across regimes reveals which reasoning-relevant features are reinforced, weakened, or remain stable.

\subsection{Experimental Design for Difficulty-Wise Feature Dynamics}
\label{sec:experiment}

\begin{wrapfigure}{r}{0.6\textwidth}
    \vspace{-10pt}
    \centering
    \includegraphics[width=1\linewidth]{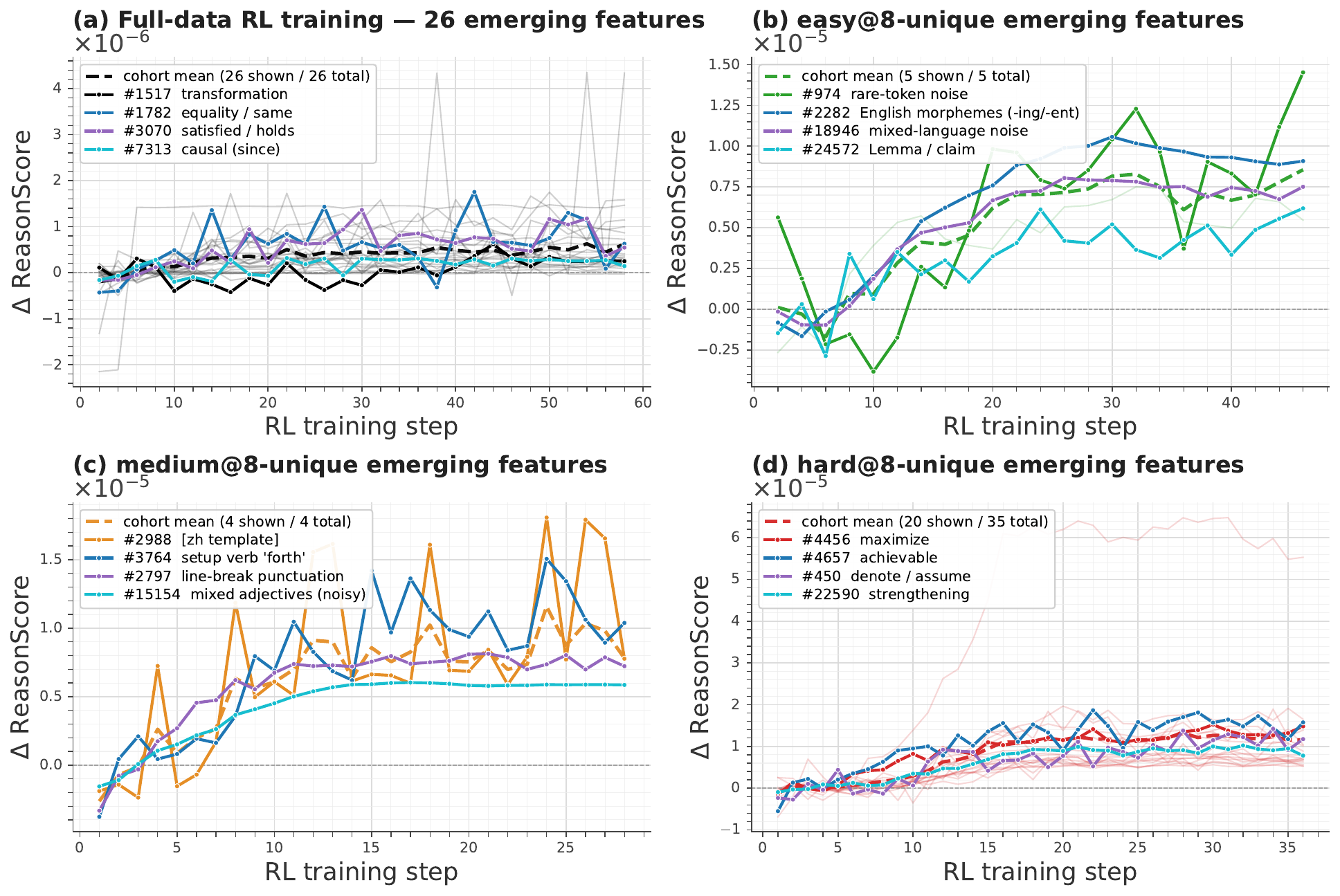}
    \caption{Per-split emergence of new reasoning features. Each panel plots the relative-to-early-mean ReasonScore trajectories of the emerging T-SAE features in one regime.}
    \label{fig:emerging_features_2x2}
    \vspace{-15pt}
\end{wrapfigure} 

In our experiments, following~\citep{galichin2026have}, we study feature dynamics in the model's intermediate residual stream. We train a T-SAE probe on the layer-16 residual stream of the full-data RL model at the final training step $\tau=58$, and then keep the probe frozen for all analyses. This gives a shared feature basis for comparing different RL-trained models. We apply the same frozen T-SAE to models trained on \texttt{easy}@8, \texttt{medium}@8, and \texttt{hard}@8 subsets, and compare their feature suppression and reinforcement patterns with the full-data model. Appendix~\ref{app:sae_tsae_comparison} provides a controlled SAE and T-SAE comparison at the same training step, supporting our use of T-SAE as the feature probe.

\textbf{Difficulty-Wise Feature Dynamics.} We first examine emerging reasoning features across full-data and difficulty-controlled training.  A feature is considered \emph{$X$-unique emerging} if its \textsc{ReasonScore} increases strongly under split $X$ while remaining flat or decreasing under the other difficulty splits. Concretely, we require Spearman $r>0.55$ over training steps and a top-$0.5\%$ late-vs.-early gain on split $X$, together with $r<0.30$ and at least $2\times$ smaller gain on the other two splits.  This criterion allows us to separate features that emerge as part of general RLVR training from those specifically recruited by easy, medium, or hard samples.

\begin{figure}[htbp]
    \centering
    \includegraphics[width=1.0\linewidth]{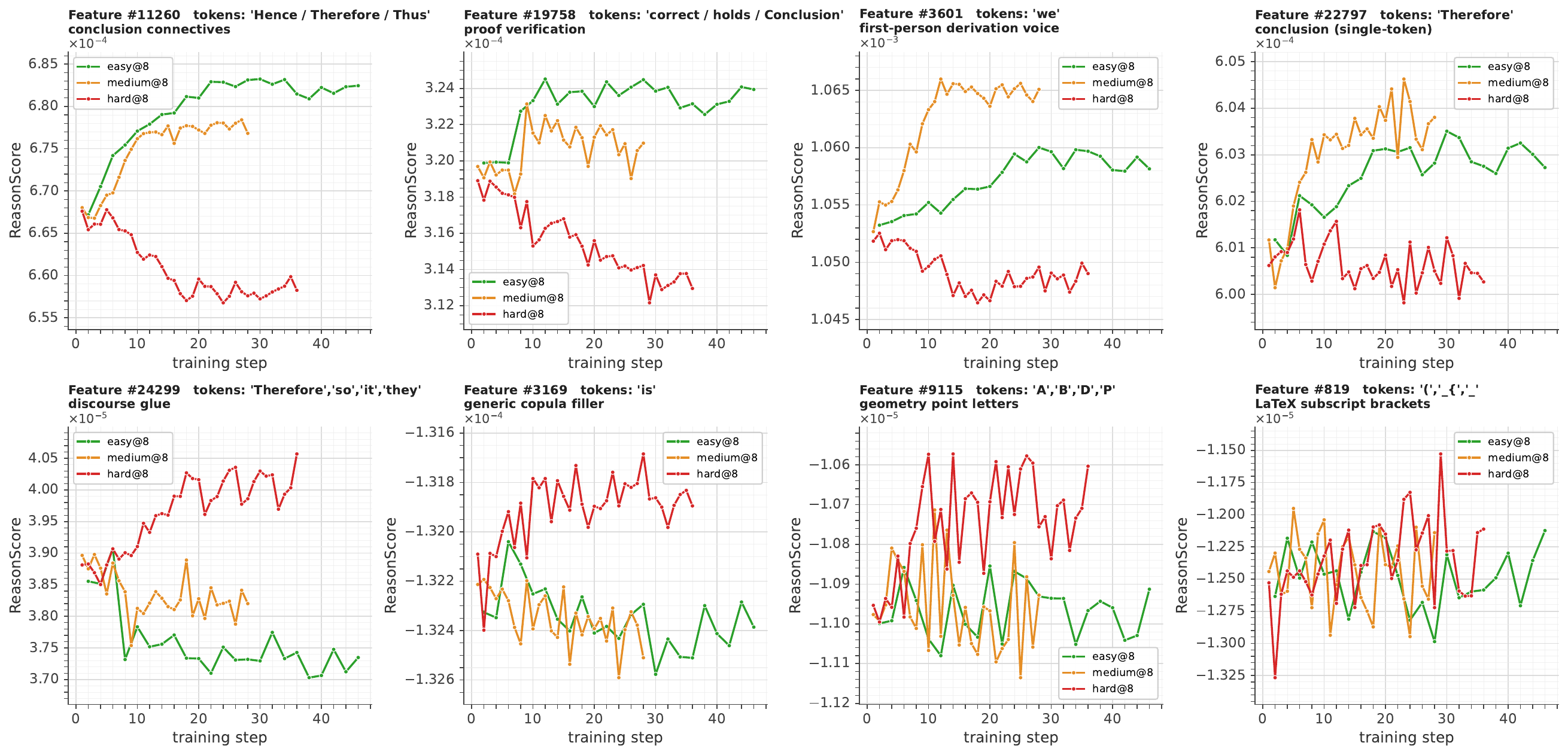}
    \caption{Token-level T-SAE feature dynamics along representative reasoning trajectories after RL on samples of different difficulty. The top row visualizes features suppressed by \texttt{Hard}@8 training, and the bottom row visualizes features reinforced by \texttt{Hard}@8 training. Hard samples tend to weaken existing reasoning features while amplifying more local or surface-level patterns.}
    \label{fig:tsae_feature}
    \vspace{-15pt}
\end{figure}

As shown in Figure~\ref{fig:emerging_features_2x2}, the resulting feature counts reveal a strong imbalance. While \texttt{easy}@8 induces only $5$ unique emerging features and \texttt{medium}@8 induces $4$, \texttt{hard}@8 induces $35$ unique emerging features, substantially more than the other regimes. Moreover, the hard-unique cohort shows larger and more synchronized positive shifts in \textsc{ReasonScore}, suggesting that hard samples recruit features that easier samples do not activate. Thus, hard samples are mechanistically rich: they do not merely strengthen existing reasoning circuitry, but can induce qualitatively new reasoning-associated features. We provide a semantic inspection of representative emerging features in Appendix~\ref{app:semantic_emerging}.

The token-level trajectories in Figure~\ref{fig:tsae_feature} clarify this difference. 
\texttt{medium}@8 consistently strengthens interpretable reasoning markers such as conclusion connectives, proof verification tokens, first-person derivation markers, and explicit conclusion tokens. In contrast, \texttt{easy}@8 more strongly activates features associated with direct calculation and concise answer production, leading to shorter responses. See Figure~\ref{fig:tsae_dynamics_per_split} for more details in Appendix~\ref{app:one_dynamics}. \texttt{hard}@8 often weakens reasoning-related features while reinforcing more local or surface-form patterns, such as discourse glue, copula fillers, geometry point letters, and \LaTeX{} subscript tokens. Thus, easy samples tend to compress reasoning into direct solution patterns, medium samples reinforce transferable reasoning routines, and hard samples may amplify narrow heuristics instead of robust reasoning.

These results explain why hard samples have inconsistent effects: they can reinforce some internal features, but this reinforcement is often accompanied by broad suppression of existing reasoning features. Appendix~\ref{app:one_dynamics} further confirms this pattern at the one-sample level: beneficial hard samples tend to reinforce reasoning-related features, whereas harmful hard samples suppress existing reasoning features or amplify surface-level patterns. 
Consequently, hard samples may occasionally improve performance but more often destabilize or degrade the model's original capabilities.

\section{Leveraging Mechanistic Insights for Better Training}

Under the T-SAE feature-dynamics framework, hard samples are not inherently uninformative; their forward-solving form often hides learnable signals from the current policy.  We therefore introduce two lightweight difficulty-adaptive interventions, backward-reasoning rewrites (Section~\ref{sec:rewrite_hard_samples}) and T-SAE-guided training signals (Section~\ref{sec:rfgo}), to validate the functional role of the reasoning features identified above.  Our goal is not to provide a complete algorithmic search for state-of-the-art RLVR performance, but to test whether the discovered feature mechanisms can be used to improve reward exposure and credit assignment for hard samples.

\subsection{Rewriting Hard Samples via Backward Reasoning}
\label{sec:rewrite_hard_samples}

As a simple data-side intervention, we convert \texttt{hard}@8 samples into backward-reasoning variants. Given a hard sample with question $q$ and answer $a$, we enumerate the numerical quantities in $q$. For each quantity, inspired by~\cite{yu2024metamath}, we replace its value with an unknown variable $z$ and create two inverse templates: one inserts the original answer as a condition and asks the model to recover $z$; the other keeps the masked question in its original form, appends “\texttt{If we know the answer to the above question is $\cdot$},” and asks for $z$. A sample with $m$ numerical quantities therefore yields up to $2m$ rewritten questions. These rewrites preserve the mathematical relation in the original prompt while converting an unreachable forward problem into more tractable inverse reasoning tasks. Example\hyperref[ex:rewrite_hard_sample]{~4.1} in Appendix~\ref{appendix:rewrite_example} shows one instance.

\textbf{Results.} We construct \texttt{AugHard}@8 by applying both templates to every numerical quantity in the \texttt{hard}@8 subset. Table~\ref{tab:rewrite_hard_samples} shows that \texttt{AugHard}@8 consistently improves over direct \texttt{hard}@8 training and surpasses the full-data baseline on AMC, AIME-2024, and MinervaMath. This result supports the view that hard samples can be useful once their reward-sparse forward form is converted into a form closer to the policy's current solvability frontier.
\begin{table}[htbp!]    
    \vspace{-10pt}
    \centering
    \small
    \setlength{\tabcolsep}{4pt}
    \caption{Performance comparison between training on the Full dataset, the original \texttt{hard}@8 subset, and the rewritten \texttt{AugHard}@8 subset.}
    \label{tab:rewrite_hard_samples}
    \resizebox{0.82\linewidth}{!}{
    \begin{tabular}{lccccc}
        \toprule
        Method & AMC & MATH-500 & AIME-2024 & Minervamath & Olympiadbench \\
        \midrule
        Full data & 40.96 & \textbf{72.20} & 10.00 & \underline{25.37} & \textbf{34.82} \\
        \texttt{hard}@8 & 39.76 & 65.00 & 3.33 & 15.81 & 30.65 \\
        \texttt{AugHard}@8 & \underline{43.37} & 68.40 & \underline{13.33} & \textbf{26.47} & 32.74 \\
        RFGO & \textbf{44.56} & \underline{72.00} & \textbf{16.67} & 25.00 & \underline{33.78} \\
        \bottomrule
    \end{tabular}
    }
    \vspace{-20pt}
\end{table}

\subsection{Reasoning Feature-Guided Optimization}
\label{sec:rfgo}

Different difficulty regimes induce distinct trajectories in the T-SAE reasoning-feature space. 
We use this signal to guide RLVR through \textbf{Reasoning Feature-Guided Optimization} (RFGO), which has two components: token-level weighting for normal verifier-supervised updates and proxy rewards for zero-variance rollout groups.

\textbf{RFGO Token Weighting.}
RFGO weights each token by the future movement it induces in the reasoning-feature space. 
Given a response split into sentence spans $\{\boldsymbol{y}_k\}_{k=1}^{K}$, we first compute the span-level feature divergence as $D_k=\|\bar{\boldsymbol{z}}^{\theta}_{S_k,\mathcal{F}_{\mathrm{reason}}}-\bar{\boldsymbol{z}}^{\mathrm{ref}}_{\boldsymbol{y}_k,\mathcal{F}_{\mathrm{reason}}}\|_1$, where $\bar{\boldsymbol{z}}_{\boldsymbol{y}_k}$ denotes the average T-SAE activation over tokens in span $\boldsymbol{y}_k$, and $\mathcal{F}_{\mathrm{reason}}$ is the set of reasoning-associated features. 
After mean-centering $D_k$ within the rollout as $\widetilde{D}_k$, the RFGO signal for token $t\in \boldsymbol{y}_k$ is $\mathrm{RFGO}_t=\sum_{r=k}^{K}\gamma^{r-k}\widetilde{D}_r$. 
We convert it into a clipped influence weight $w_t=\texttt{clip}(\exp(\beta\,\mathrm{RFGO}_t),1-\epsilon_w,1+\epsilon_w)$. 
The verifier advantage still determines the update direction, while $w_t$ adjusts the token-level update magnitude according to future reasoning-feature movement.

\textbf{RFGO Proxy Reward.}
For some samples, all rollouts in a GRPO group receive the same verifier reward, i.e., $\{R_i\}_{i=1}^{G}$ is all-zero or all-one. 
Such zero-variance groups produce no informative group-relative advantage. 
RFGO therefore uses a feature-based proxy reward only in this case. 
For each rollout $\boldsymbol{y}_i$, we compute $R_i^{\mathrm{RFGO}}=\frac{\sum_t M_{i,t}\sum_{j\in\mathcal{F}_{\mathrm{reason}}}\omega_j z^\theta_{i,t,j}}{\sum_t M_{i,t}}$, where $M_{i,t}$ masks valid response tokens and $\omega_j$ is the feature weight derived from \textsc{ReasonScore}. 
We then set $\widetilde{R}_i=R_i^{\mathrm{RFGO}}$ if $R_1=\cdots=R_G$, and otherwise keep the original verifier reward, $\widetilde{R}_i=R_i$. 
Thus, RFGO preserves binary verifier rewards when they are informative, but introduces a reasoning-feature-based ranking when verifier rewards have zero variance.

\textbf{Results.} RFGO achieves the best overall performance without rewriting the training data. 
It obtains the highest score on AMC and AIME-2024, nearly matches Full data on MATH-500, and improves the average score from $30.91$ for \texttt{Hard}@8 and $36.67$ for Full data to $38.40$. 
These results indicate that reasoning-feature guidance can better exploit difficult samples by preserving useful verifier feedback while reducing the instability caused by uninformative or misleading hard-sample rewards.

\section{Conclusion}

In this work, we presented a mechanistic study of how samples across the full difficulty spectrum shape RLVR training for LLMs. Through controlled experiments and one-sample amplification, we demonstrated that hard samples produce sparse, low-quality learning signals that destabilize optimization and degrade performance, while medium-difficulty samples yield the strongest data efficiency. Critically, we showed that sample informativeness is not a static property but depends on the dynamic interaction between task difficulty and the model's evolving capability. Moving beyond reward-level analysis, we employed Temporal Sparse Autoencoders (T-SAEs) to track the evolution of sparse feature activations throughout RL training, revealing that samples of different difficulty levels selectively reinforce or suppress distinct reasoning feature dynamics that are invisible from advantage signals alone. Guided by these mechanistic insights, we proposed two complementary interventions: a backward reasoning reformulation that converts hard samples into learnable inverse problems, and a T-SAE feature-based training signal that supports both token-level weighting and direct hard-sample rewards. Experiments on multiple mathematical reasoning benchmarks confirmed that these strategies significantly outperform direct training on hard subsets, with the rewritten samples surpassing full-data baselines on several benchmarks.

\clearpage

\bibliography{ref}

\newpage

\appendix

\section*{Appendix}

\section{Related Work}
\label{relatedwork}
\paragraph{Reinforcement Learning with Verifiable Reward.} Reinforcement Learning from Verifiable Rewards has been proposed as a promising framework for enhancing the reasoning capabilities of large language models by leveraging rule-based or verifiable functions. Early RLVR pipelines typically built on Proximal Policy Optimization (PPO)~\citep{schulman2017proximal}, which provides stable updates but incurs high computational cost and sensitivity due to its critic model. VAPO~\citep{yue2025vapo} revisits this value-based line for long-CoT reasoning by improving the reliability and efficiency of PPO-style training. To reduce the overhead of value-model learning, GRPO~\citep{shao2024deepseekmath} replaces the critic with group-relative reward normalization, while Dr.\ GRPO~\citep{liu2025understanding} analyzes and corrects biases in R1-zero-like training. DAPO~\citep{yu2025dapo} further improves long-CoT RLVR through decoupled clipping, dynamic sampling, token-level loss, and overlong reward shaping; GSPO~\citep{zheng2025group} moves clipping and optimization from the token level to the sequence level; and DCPO~\citep{yang2025dcpo} dynamically adjusts clipping to stabilize policy updates. In parallel, ReMax~\citep{li2023remax}, REINFORCE++~\citep{hu2025reinforce++}, and RLOO~\citep{ahmadian2024back} simplify RLVR optimization with REINFORCE-style estimators and lightweight baselines. Orthogonal to optimizer design, PRIME~\citep{cui2025process} introduces implicit process rewards, Scaf-GRPO~\citep{zhang2026scafgrpo} provides scaffolded guidance for zero-reward hard problems, ExGRPO~\citep{zhan2026exgrpo} improves sample efficiency through experience reuse, and NPO~\citep{qin2026near} uses near-future checkpoints from the same training run as mixed-policy trajectory sources to balance trajectory quality and distributional closeness. Beyond language reasoning, DanceGRPO~\citep{xue2025dancegrpo} and Flow-GRPO~\citep{liu2025flowgrpo} extend GRPO-style online RL to visual generation and flow matching, respectively.These works ask how to make RLVR optimization more stable, efficient, or scalable. In contrast, we study which samples provide useful RLVR signals and why hard samples are unstable.

\paragraph{Sample Difficulty and Effective in RLVR.} Beyond optimizer design, recent work studies how sample difficulty determines whether a prompt yields useful RLVR signal.~\citep{yue2025does} shows that RL gains are often bounded by the capabilities already present in the base model, suggesting that prompts outside the model's reachable capability frontier may provide weak or unreliable feedback. Following the broader curriculum-learning principle~\citep{bengio2009curriculum}, E2H Reasoner~\citep{parashar2025curriculum} schedules training from easy to hard tasks, while Online Difficulty Filtering~\citep{bae2026online} filters prompts online to emphasize intermediate-difficulty examples that are neither already saturated nor consistently failed. VCRL~\citep{jiang2025vcrl} treats rollout reward variance as a difficulty proxy, prioritizing prompts whose mixture of successful and failed samples can produce nonzero group-relative advantages. DEPO~\citep{tang2026highefficiency} improves data efficiency by combining prior data selection with online exploration and replay, thereby allocating rollout budget to prompts that are more likely to yield informative updates. At the low-signal end of the difficulty spectrum, RL-ZVP~\citep{le2026noprompt} revisits zero-variance prompts and applies entropy-guided advantage shaping, suggesting that samples with vanishing advantages can still affect exploration when their token-level uncertainty is used. DARS and DARS-B~\citep{yang2026deepsyn} adapt exploration along depth and breadth to expand the solvable region, QuestA~\citep{li2026questa} augments questions to elicit harder reasoning behaviors, ReGFT~\citep{wu2026learnhard} uses reference-guided fine-tuning to make hard problems more learnable during RL, and P$^2$O~\citep{lu2026p2o} jointly optimizes prompts and policies so that the training distribution evolves with the learner. However, our work is complementary but focuses on a different question: why do hard samples behave unstably in RLVR? 
Rather than only selecting, augmenting, or exploring hard samples, we study their effects through curriculum-level, one-sample, and T-SAE feature-level analyses.  We show that hard samples are heterogeneous: some induce useful policy changes, but many suppress existing reasoning features or amplify brittle shortcuts, thereby damaging downstream performance.

\paragraph{Sparse Autoencoders for LLM Interpretability.} Sparse Autoencoders~\citep{ng2011sparse} provide an unsupervised tool for decomposing dense language-model activations into sparse latent features, offering a practical route to study superposition and polysemanticity in LLM representations.~\citep{huben2024sparse} shows that SAE features in language models are more interpretable and monosemantic than raw neurons or arbitrary activation directions, and can support causal interventions on model behavior.~\citep{gao2025scaling} further improves SAE training with controlled sparsity and systematic feature-quality metrics, demonstrating that larger SAEs recover richer features and can be trained even on frontier-model activations. Gemma Scope~\citep{lieberum2024gemma} lowers the barrier for layer-wise and sub-layer-wise analysis of modern open models, while the automated interpretability pipeline~\citep{paulo2025automatically} scales natural-language interpretation to millions of SAE features and introduces intervention scoring to evaluate feature effects under intervention. Beyond feature discovery, Sparse Feature Circuits~\citep{marks2025sparse} connects SAE features into causal graphs for explaining and editing model behaviors, and AbsTopK~\citep{zhu2026abstopk} revisits the sparse-coding formulation to recover bidirectional conceptual features. For reasoning models, ReasonScore~\citep{galichin2026have} identifies SAE features associated with uncertainty, exploratory thinking, and reflection in DeepSeek-R1-style models, providing direct evidence that sparse features can expose internal reasoning mechanisms. Prior SAE work primarily studies feature discovery, monosemanticity, causal intervention, automated interpretation, and sparse feature circuits in pretrained or post-trained language models. For reasoning models, ReasonScore further shows that SAE features can capture uncertainty, exploration, and reflection-related behaviors. 
Our work is complementary but focuses on RLVR dynamics: we use T-SAE to analyze how samples of different difficulty reinforce or suppress reasoning features during policy optimization. 

\section{More About Policy Algorithms and Definitions}
\label{appendix:core_alg}

\subsection{More About Policy Optimization Algorithms}

\paragraph{Group Relative Policy Optimization (GRPO).} GRPO~\citep{shao2024deepseekmath} is an efficient RL algorithm introduced by DeepSeek for training its \texttt{DeepSeek-Math} model, aiming to reduce the training cost of PPO-style~\citep{schulman2017proximal} reasoning. GRPO removes the value network by estimating advantages from the in-group standardized reward. Concretely, for each query–answer pair $(\boldsymbol{x},\boldsymbol{y})$, the rollout policy $\pi_{\theta_{\text{old}}}$ samples a group of responses $\{o^{i}\}_{i=1}^{G}$ with binary outcome rewards $\{R^{i}\}_{i=1}^{G}$ defined by a verifier. Let
\begin{equation}
    \hat{A}^i_t = \frac{R^i - \text{mean}(\{R^i\}_{i=1}^G)}{\text{std}(\{R^i\}_{i=1}^G)}, 
    \quad 
    \text{where} \quad 
    R^i = 
    \begin{cases}
        1 & \text{if } \texttt{is\_equivalent}(a, o^i), \\[6pt]
        0 & \text{otherwise}.
    \end{cases}
    \label{eq:grpo_adv}
\end{equation}
GRPO then optimizes a PPO-style clipped objective with an additional KL penalty to a reference model
\begin{equation}
\begin{aligned}
    \mathcal{J}_{GRPO}(\theta) = &\mathbb{E}_{(q,a) \sim \mathcal{D}, \{o^{i}\}_{i=1}^{G} \sim \pi_{\theta\text{old}}(\cdot \mid q)}\\
    &\frac{1}{G} \sum_{i=1}^G \frac{1}{| o^i |} \sum_{t=1}^{| o^i |} \bigg[\min \Big(r^i_t(\theta) \hat{A}^i_t, \text{clip}(r^i_t(\theta), 1-\epsilon, 1+\epsilon)\hat{A}^i_t\Big)- \beta \mathcal{D}_{KL}(\pi_\theta \| \pi_{\text{ref}}) \bigg],
\end{aligned}
    \label{eq:grpo_obj}
\end{equation}
where the per-token likelihood ratio is
\begin{equation}
    r^i_t(\theta) = \frac{\pi_\theta(o^i_t \mid q,o^i_{<t})}{\pi_{\theta\text{old}}(o^i_t \mid q,o^i_{<t})}.
\end{equation}
$G\in\mathbb{N}$ is the group size, $\epsilon>0$ is the PPO clipping parameter, $\beta\ge 0$ controls KL regularization, and $\varepsilon \ll 1$ stabilizes the standard deviation.

\subsection{More About Temporal Sparse Autoencoders.}

Sparse Autoencoders (SAEs) decompose dense model activations into sparse latent features. 
Given an activation $\boldsymbol{h}_t\in\mathbb{R}^d$ at token position $t$, an SAE encodes it into an overcomplete sparse code and reconstructs the original activation:
\begin{equation}
    \boldsymbol{z}_t
    =
    \zeta(\boldsymbol{h}_t)
    =
    \mathrm{TopK}\!\left(
        \sigma\!\left(
        \boldsymbol{W}^{\mathrm{enc}}\boldsymbol{h}_t
        +
        \boldsymbol{b}^{\mathrm{enc}}
        \right)
    \right),
    \qquad
    \hat{\boldsymbol{h}}_t
    =
    \xi(\boldsymbol{z}_t)
    =
    \boldsymbol{W}^{\mathrm{dec}}\boldsymbol{z}_t
    +
    \boldsymbol{b}^{\mathrm{dec}},
\end{equation}
where $\boldsymbol{W}^{\mathrm{enc}}\in\mathbb{R}^{m\times d}$, 
$\boldsymbol{W}^{\mathrm{dec}}\in\mathbb{R}^{d\times m}$, 
$\boldsymbol{b}^{\mathrm{enc}}\in\mathbb{R}^{m}$, 
$\boldsymbol{b}^{\mathrm{dec}}\in\mathbb{R}^{d}$, and $m\gg d$. 
Each coordinate of $\boldsymbol{z}_t$ corresponds to a sparse latent feature, whose value indicates the activation strength of that feature at token position $t$.

Temporal Sparse Autoencoders (T-SAEs) extend standard SAEs by incorporating the sequential structure of language. 
The key assumption is that high-level semantic features should remain relatively stable across nearby tokens in the same sequence, while low-level syntactic or token-specific features may fluctuate more rapidly. 
To model this distinction, T-SAE partitions the sparse code into high-level and low-level components:
\begin{equation}
    \boldsymbol{z}_t
    =
    \big[
    \boldsymbol{z}^{0:h}_t,
    \boldsymbol{z}^{h:m}_t
    \big],
\end{equation}
where $\boldsymbol{z}^{0:h}_t\in\mathbb{R}^{h}$ denotes the first $h$ high-level features and $\boldsymbol{z}^{h:m}_t\in\mathbb{R}^{m-h}$ denotes the remaining low-level features. 
Correspondingly, the decoder matrix is partitioned as
\begin{equation}
    \boldsymbol{W}^{\mathrm{dec}}
    =
    \big[
    \boldsymbol{W}^{\mathrm{dec}}_{0:h},
    \boldsymbol{W}^{\mathrm{dec}}_{h:m}
    \big],
\end{equation}
where $\boldsymbol{W}^{\mathrm{dec}}_{0:h}\in\mathbb{R}^{d\times h}$ and 
$\boldsymbol{W}^{\mathrm{dec}}_{h:m}\in\mathbb{R}^{d\times(m-h)}$.

The reconstruction objective contains a high-level reconstruction term and a full reconstruction term:
\begin{equation}
    \mathcal{L}_{\mathrm{rec}}
    =
    \left\|
    \boldsymbol{h}_t
    -
    \left(
    \boldsymbol{W}^{\mathrm{dec}}_{0:h}
    \boldsymbol{z}^{0:h}_t
    +
    \boldsymbol{b}^{\mathrm{dec}}
    \right)
    \right\|_2^2
    +
    \left\|
    \boldsymbol{h}_t
    -
    \left(
    \boldsymbol{W}^{\mathrm{dec}}
    \boldsymbol{z}_t
    +
    \boldsymbol{b}^{\mathrm{dec}}
    \right)
    \right\|_2^2 .
    \label{eq:tsae_rec_loss}
\end{equation}
The first term encourages high-level features to capture stable semantic structure, while the second term preserves full reconstruction quality using both high- and low-level features.

To encourage temporal consistency, T-SAE further adds a contrastive objective on the high-level features. 
For paired activations $(\boldsymbol{h}^{(i)}_t,\boldsymbol{h}^{(i)}_{t-1})$ from the same sequence, let
\begin{equation}
    \boldsymbol{u}^{(i)}_t
    =
    \boldsymbol{z}^{0:h}(\boldsymbol{h}^{(i)}_t),
    \qquad
    \boldsymbol{u}^{(i)}_{t-1}
    =
    \boldsymbol{z}^{0:h}(\boldsymbol{h}^{(i)}_{t-1}),
\end{equation}
and let $s(\cdot,\cdot)$ denote cosine similarity. 
The contrastive loss is
\begin{equation}
    \mathcal{L}_{\mathrm{contr}}
    =
    -
    \frac{1}{N}
    \sum_{i=1}^{N}
    \log
    \frac{
    \exp\!\left(s(\boldsymbol{u}^{(i)}_t,\boldsymbol{u}^{(i)}_{t-1})\right)
    }{
    \sum_{j=1}^{N}
    \exp\!\left(s(\boldsymbol{u}^{(i)}_t,\boldsymbol{u}^{(j)}_{t-1})\right)
    } .
    \label{eq:tsae_contrastive_loss}
\end{equation}
This objective pulls together high-level features from nearby tokens in the same sequence while pushing apart high-level features from different examples, preventing collapse to constant representations.

The full T-SAE objective is
\begin{equation}
    \mathcal{L}_{\mathrm{TSAE}}
    =
    \mathcal{L}_{\mathrm{rec}}
    +
    \lambda_{\mathrm{contr}}
    \mathcal{L}_{\mathrm{contr}} .
    \label{eq:tsae_full_loss}
\end{equation}
In our analysis, this temporal structure is important because reasoning unfolds across multiple generated tokens rather than at isolated positions. 
T-SAE therefore provides a feature basis better suited for tracking reasoning-feature dynamics during RLVR.

\subsection{More About Sample Difficulty}
\label{app:task_difficulty}

\paragraph{Low-Variance \texttt{pass}@$k$ Estimation.} Directly computing \texttt{pass}@$k$ using only $k$ sampled outputs per problem can lead to high variance. To mitigate this, we follow the unbiased estimation method proposed by Chen \textit{et al}. Specifically, for each problem $\boldsymbol{x}_i$ from the evaluation dataset $\mathcal{D}$, we generate $n$ samples ($n \geq k$) and count the number of correct samples as $c_i$. The unbiased estimator of \texttt{pass}@$k$ over the dataset is given by: 
\begin{equation}
    \texttt{pass}@k := \mathbb{E}_{\boldsymbol{x}_i \in \mathcal{D}} \left[1-\frac{\binom{n-c_i}{k}}{\binom{n}{k}}\right]
\end{equation}
With this formulation, we can easily estimate \texttt{pass}@$k$ with low variance across all $k \leq n$.

\paragraph{Sample Difficulty.} Following~\cite{yue2025does}, we use \texttt{pass}@$k$ as a proxy for sample difficulty, where $k$ is set to the group size used in GRPO. In this paper, we set group size is equal to 8. We then define a problem as \textbf{\textit{unsolved within $k$ attempts}} if none of the $k$ candidates passes verification, \textit{i.e.}, $\texttt{pass}@k = 0$. Such a problem is categorized as a difficulty-$k$ problem, which we refer to as a \texttt{hard}@$k$ sample for convenience. Conversely, to identify easy samples, we introduce \texttt{easy}@$k$. A problem is categorized as an \texttt{easy}@$k$ sample if all of its $k$ sampled candidates pass verification, indicating that the model can solve the problem consistently across rollouts. To further characterize intermediate difficulty, we define \texttt{medium}@$k$ samples as those satisfying $\texttt{pass}@k \neq 0$ and not belonging to easy@$k$, i.e., samples where at least one but not all rollouts pass verification.

\section{Additional Examples of Harmful Hard Samples}
\label{appendix:failure_examples}

We provide additional representative examples of harmful hard samples. Example\hyperref[ex:goal]{~2.2} shows misunderstanding of the target goal, where the model treats a make-up requirement as a simple cumulative-goal problem. Example\hyperref[ex:mismatch]{~2.3} illustrates a question--answer mismatch, where the provided label corresponds to the daily cost rather than the total cost requested by the question. Example\hyperref[ex:ops]{~2.4} demonstrates missing aggregation operations in the final answer. Example\hyperref[ex:py]{~2.5} shows a Python calculation error that contradicts the correct arithmetic. Example\hyperref[ex:type]{~2.6} captures inconsistent output value types, where the model alternates between integer and decimal monetary formats. Example\hyperref[ex:logic]{~2.7} presents a logical error in combining multiple constraints.

\phantomsection\label{ex:goal}
\begin{paperlistingexample}{Example 2.2: Misunderstanding the Goal}
(*@\textbf{Question:}@*) Shawna's workout goal is 30 situps. On Monday, Shawna was only able to do 12 situps, so she decided that she would make up for the rest on Tuesday. However, she was only able to do 19 situps on Tuesday. How many situps would Shawna have to do on Wednesday to meet her minimum goal and make up for the ones she didn't do? ((*@\textbf{Answer:}@*) (*@\textcolor{red!70!black}{59}@*))
(*@\textbf{Response:}@*) (*@\color{red!70!black}@*)Shawna's goal is to do 30 situps in total.(*@\color{black}@*) On Monday, she did 12 situps, and on Tuesday, she did 19 situps. (*@\color{red!70!black}@*)To find out how many situps she needs to do on Wednesday to meet her goal, we can calculate the total number of situps she has already done and then subtract that from her goal.(*@\color{black}@*)

1. Calculate the total number of situps Shawna has done so far:
\[ \text{Total situps done so far} = 12 + 19 = 31 \]

2. (*@\color{red!70!black}@*)Since Shawna's goal is 30 situps, she has already exceeded her goal by 1 situp. Therefore, she doesn't need to do any more situps on Wednesday to meet her goal. However, if we interpret the problem as finding out how many situps she would need to do on Wednesday to make up for the ones she didn't do (which is not possible since she has already exceeded her goal), we can still calculate it.(*@\color{black}@*)

3. The number of situps Shawna needs to do on Wednesday to make up for the ones she didn't do is:
\[ (*@\color{red!70!black}@*)\text{Situps needed on Wednesday} = 30 - (12 + 19) = 30 - 31 = -1(*@\color{black}@*) \]

(*@\color{red!70!black}@*)Since it's not possible to do a negative number of situps, the answer is 0.(*@\color{black}@*) Let's confirm this with Python code.
```python
# Shawna's goal
(*@\color{red!70!black}@*)goal = 30(*@\color{black}@*)

# Situps done on Monday and Tuesday
situps_monday = 12
situps_tuesday = 19

# Total situps done so far
total_done = situps_monday + situps_tuesday

# Situps needed on Wednesday to meet the goal
(*@\color{red!70!black}@*)situps_needed_wednesday = goal - total_done(*@\color{black}@*)

# Since it's not possible to do a negative number of situps, we set it to 0
(*@\color{red!70!black}@*)if situps_needed_wednesday < 0:
    situps_needed_wednesday = 0(*@\color{black}@*)

# Output the final answer
print(f"\boxed{{{situps_needed_wednesday}}}")
```
```output
(*@\color{red!70!black}@*)\boxed{0}(*@\color{black}@*)
```
The final answer is \(\boxed{(*@\color{red!70!black}@*)0(*@\color{black}@*)}\).
\end{paperlistingexample}

\phantomsection\label{ex:mismatch}
\begin{paperlistingexample}{Example 2.3: \textit{Question Mismatches Answer}}
(*@\textbf{Question:}@*) Henry took 9 pills a day for 14 days. Of these 9 pills, 4 pills cost $1.50 each, and the other pills each cost $5.50 more. How much did he spend in total on the pills? ((*@\textbf{Answer:}@*) (*@\textcolor{red!70!black}{41}@*))
(*@\textbf{Response:}@*) First, let's break down the problem:

1. Henry takes 9 pills a day for 14 days, so the total number of pills he takes is \(9 \times 14 = 126\).
2. Out of these 9 pills, 4 pills cost $1.50 each, and the other 5 pills cost $1.50 + $5.50 = $7.00 each.
3. The total cost for the 4 pills is \(4 \times 1.50 = 6.00\).
4. The total cost for the 5 pills is \(5 \times 7.00 = 35.00\).
5. The total cost for the 9 pills is \(6.00 + 35.00 = 41.00\).
6. (*@\color{red!70!black}@*)Since Henry takes these 9 pills every day for 14 days, the total cost for all the pills is \(41.00 \times 14 = 574.00\).(*@\color{black}@*)

Now, let's write the Python code to verify this calculation.
```python
# Number of pills per day
pills_per_day = 9

# Number of days
days = 14

# Cost of 4 pills
cost_4_pills = 4 * 1.50

# Cost of 5 pills
cost_5_pills = 5 * (1.50 + 5.50)

# Total cost for 9 pills
total_cost_9_pills = cost_4_pills + cost_5_pills

# Total cost for all pills over 14 days
(*@\color{red!70!black}@*)total_cost = total_cost_9_pills * days(*@\color{black}@*)

# Output the final answer
print(f"\boxed{{{int(total_cost)}}}")
```
```output
(*@\color{red!70!black}@*)\boxed{574}(*@\color{black}@*)
```
The total amount Henry spent on the pills is \(\boxed{(*@\color{red!70!black}@*)574(*@\color{black}@*)}\).
\end{paperlistingexample}

\phantomsection\label{ex:ops}
\begin{paperlistingexample}{Example 2.4: Missing Multiplication or Addition}
(*@\textbf{Question:}@*) In a section of the forest, there are 100 weasels and 50 rabbits. Three foxes invade this region and hunt the rodents. Each fox catches an average of 4 weasels and 2 rabbits per week. How many rabbits and weasels will be left after 3 weeks? ((*@\textbf{Answer:}@*) (*@\textcolor{red!70!black}{96}@*))
(*@\textbf{Response:}@*) 1. First, we need to calculate the number of weasels and rabbits caught by the foxes in one week.
2. Each fox catches 4 weasels and 2 rabbits per week. With 3 foxes, the total number of weasels caught per week is \(3 \times 4 = 12\) weasels, and the total number of rabbits caught per week is \(3 \times 2 = 6\) rabbits.
3. Over 3 weeks, the total number of weasels caught is \(12 \times 3 = 36\) weasels, and the total number of rabbits caught is \(6 \times 3 = 18\) rabbits.
4. The initial number of weasels is 100, so after 3 weeks, the number of weasels left is \(100 - 36 = 64\).
5. The initial number of rabbits is 50, so after 3 weeks, the number of rabbits left is \(50 - 18 = 32\).
6. (*@\color{red!70!black}@*)The final answer should be boxed.(*@\color{black}@*)

Let's implement this in Python to ensure the calculations are correct.
```python
# Initial number of weasels and rabbits
initial_weasels = 100
initial_rabbits = 50

# Number of foxes and their catch rates
foxes = 3
weasels_per_fox_per_week = 4
rabbits_per_fox_per_week = 2

# Total number of weasels and rabbits caught per week
weasels_caught_per_week = foxes * weasels_per_fox_per_week
rabbits_caught_per_week = foxes * rabbits_per_fox_per_week

# Total number of weasels and rabbits caught in 3 weeks
weasels_caught_in_3_weeks = weasels_caught_per_week * 3
rabbits_caught_in_3_weeks = rabbits_caught_per_week * 3

# Number of weasels and rabbits left after 3 weeks
weasels_left = initial_weasels - weasels_caught_in_3_weeks
rabbits_left = initial_rabbits - rabbits_caught_in_3_weeks

# Final answer in boxed format
(*@\color{red!70!black}@*)final_answer = f"\boxed{{{int(weasels_left)} {int(rabbits_left)}}}"(*@\color{black}@*)
print(final_answer)
```
```output
(*@\color{red!70!black}@*)\boxed{64 32}(*@\color{black}@*)
```
The final answer is:

\[
(*@\color{red!70!black}@*)\boxed{64 32}(*@\color{black}@*)
\]
\end{paperlistingexample}

\phantomsection\label{ex:py}
\begin{paperlistingexample}{Example 2.5: Python Calculation Error}
(*@\textbf{Question:}@*) A cake has 8 slices and each slice contains 347 calories.  A pan of brownies has 6 brownies and each slice contains 375 calories.  How many more calories does the cake have? ((*@\textbf{Answer:}@*) (*@\textcolor{red!70!black}{526}@*))
(*@\textbf{Response:}@*) First, we need to calculate the total number of calories in the cake and the total number of calories in the pan of brownies. Then, we will find the difference between these two values.

1. Calculate the total number of calories in the cake:
\[ \text{Total calories in cake} = 8 \text{ slices} \times 347 \text{ calories/slice} \]

2. Calculate the total number of calories in the pan of brownies:
\[ \text{Total calories in brownies} = 6 \text{ brownies} \times 375 \text{ calories/brownie} \]

3. Find the difference between the total number of calories in the cake and the total number of calories in the pan of brownies:
\[ \text{Difference} = \text{Total calories in cake} - \text{Total calories in brownies} \]

Let's calculate this using Python code.
```python
# Calculate the total number of calories in the cake
total_calories_cake = 8 * 347

# Calculate the total number of calories in the pan of brownies
total_calories_brownies = 6 * 375

# Calculate the difference
difference = total_calories_cake - total_calories_brownies

# Output the final answer in boxed format
print(f"\boxed{{{int(difference)}}}")
```
```output
(*@\color{red!70!black}@*)\boxed{14}(*@\color{black}@*)
```
The cake has \(\boxed{(*@\color{red!70!black}@*)14(*@\color{black}@*)}\) more calories than the pan of brownies.
\end{paperlistingexample}

\phantomsection\label{ex:type}
\begin{paperlistingexample}{Example 2.6: Different Value Types}
(*@\textbf{Question:}@*) Tim buys 3 dozen eggs.  Eggs cost $.50 each.  How much did he pay for eggs? ((*@\textbf{Answer:}@*) (*@\textcolor{red!70!black}{18}@*))
(*@\textbf{Response:}@*) First, we need to determine the total number of eggs Tim bought. Since Tim buys 3 dozen eggs, and one dozen is equal to 12, the total number of eggs is \(3 \times 12 = 36\).

Next, we need to calculate the total cost of these eggs. Since each egg costs $0.50, the total cost is \((*@\color{red!70!black}@*)36 \times 0.50 = 18.00(*@\color{black}@*)\).

So, Tim paid (*@\color{red!70!black}@*)$18.00(*@\color{black}@*) for the eggs. Let's put this in the final answer format.

(*@\color{red!70!black}@*)\boxed{18.00}(*@\color{black}@*)
```python
# Calculate the total number of eggs
total_eggs = 3 * 12

# Calculate the total cost
total_cost = total_eggs * 0.50

# Output the final answer
print(f"\boxed{{{int(total_cost)}}}")
```
```output
\boxed{18}
```
The total cost of the eggs is (*@\color{red!70!black}@*)$18.00(*@\color{black}@*). Therefore, the final answer is:

\[
(*@\color{red!70!black}@*)\boxed{18.00}(*@\color{black}@*)
\]
\end{paperlistingexample}

\phantomsection\label{ex:logic}
\begin{paperlistingexample}{Example 2.7: Logical Error}
(*@\textbf{Question:}@*) Amaya scored 20 marks fewer in Maths than she scored in Arts. She also got 10 marks more in Social Studies than she got in Music. If she scored 70 in Music and scored 1/10 less in Maths, what's the total number of marks she scored in all the subjects? ((*@\textbf{Answer:}@*) (*@\textcolor{red!70!black}{296}@*))
(*@\textbf{Response:}@*) First, let's define the variables for the scores in each subject:
- Let \( A \) be the score in Arts.
- Let \( M \) be the score in Music.
- Let \( S \) be the score in Social Studies.
- Let \( T \) be the score in Maths.

From the problem statement, we know:
1. Amaya scored 20 marks fewer in Maths than she scored in Arts. So, \( T = A - 20 \).
2. She scored 10 marks more in Social Studies than she got in Music. So, \( S = M + 10 \).
3. She scored 70 in Music. So, \( M = 70 \).
4. (*@\color{red!70!black}@*)She scored \( \frac{1}{10} \) less in Maths than in Arts. So, \( T = A - \frac{1}{10}A = \frac{9}{10}A \).(*@\color{black}@*)

Using the value of \( M \), we can find \( S \):
\[ S = 70 + 10 = 80 \]

(*@\color{red!70!black}@*)Now, using the value of \( T \) and the relationship \( T = \frac{9}{10}A \), we can find \( A \):
\[ A - 20 = \frac{9}{10}A \]
\[ A - \frac{9}{10}A = 20 \]
\[ \frac{1}{10}A = 20 \]
\[ A = 200 \](*@\color{black}@*)

(*@\color{red!70!black}@*)Now we can find \( T \):
\[ T = 200 - 20 = 180 \](*@\color{black}@*)

So, the scores in each subject are:
- Arts: (*@\color{red!70!black}@*)\( A = 200 \)(*@\color{black}@*)
- Music: \( M = 70 \)
- Social Studies: \( S = 80 \)
- Maths: (*@\color{red!70!black}@*)\( T = 180 \)(*@\color{black}@*)

The total number of marks she scored in all the subjects is:
\[ (*@\color{red!70!black}@*)200 + 70 + 80 + 180 = 530(*@\color{black}@*) \]

Let's confirm this with Python code.
```python
# Given values and relationships
M = 70  # Score in Music
S = M + 10  # Score in Social Studies
(*@\color{red!70!black}@*)A = (M + 20) * (10 / 9)  # Score in Arts, derived from T = (9/10)A and T = A - 20(*@\color{black}@*)
T = A - 20  # Score in Maths

# Total marks
(*@\color{red!70!black}@*)total_marks = A + M + S + T(*@\color{black}@*)

# Output the final answer
print(f"\boxed{{{int(total_marks)}}}")
```
```output
(*@\color{red!70!black}@*)\boxed{530}(*@\color{black}@*)
```
The total number of marks Amaya scored in all the subjects is \(\boxed{(*@\color{red!70!black}@*)530(*@\color{black}@*)}\).
\end{paperlistingexample}

\section{Experimental Setup}
\label{appendix:exp}

\paragraph{Models.} Unless otherwise specified, all experiments in the main text used Qwen2.5-Math-1.5B~\citep{yang2024qwen2} as the default base model. To examine the effect of model scale and instruction tuning, we further evaluated Qwen2.5-Math-7B~\citep{yang2024qwen2} and DeepSeek-Math-7B-Instruct~\citep{shao2024deepseekmath}.

\paragraph{Training.} We implemented our reinforcement learning pipeline with the verl~\citep{sheng2024hybridflow} framework. The KL coefficient $\beta$ was set to 0.001 by default, and the training batch size and mini-batch size were set to 128 and 64, respectively. For each prompt, we sampled $G=8$ responses, with maximum prompt and response lengths of 1024 and 3072. The learning rate was set to $3\times10^{-6}$ for Qwen2.5-Math-1.5B and $1\times10^{-6}$ for Qwen2.5-Math-7B and DeepSeek-Math-7B-Instruct. Policy optimization was performed with GRPO~\citep{shao2024deepseekmath}, keeping the optimization algorithm fixed so that differences in performance could be attributed to the sampled data regimes rather than optimizer variants. All experiments are conducted on eight NVIDIA A100 80GB GPUs.

\paragraph{Dataset.} We conducted training experiments on MATH~\citep{hendrycks2021measuring}, a competition-style mathematical reasoning dataset containing 12,500 problems across algebra, geometry, number theory, combinatorics, and related topics. The dataset provides 7,500 training problems and 5,000 test problems, each paired with a natural-language solution and metadata such as problem type and difficulty.

\paragraph{Evaluation.} We reported \textit{pass@1} as the primary evaluation metric. Following common practice in mathematical reasoning evaluation, we evaluated models on AMC~\citep{li2024numinamath}, MATH-500~\citep{lightman2023lets}, AIME-2024~\citep{li2024numinamath}, Minervamath~\citep{lewkowycz2022solving}, and Olympiadbench~\citep{he2024olympiadbench}. These benchmarks collectively covered a broad spectrum of mathematical problem-solving skills, ranging from high-school level contest problems to advanced Olympiad-style challenges.

\paragraph{T-SAE Details.} 
We use the same frozen T-SAE from Section~\ref{sec:sae_analysis} (layer~16, $m=24{,}576$ features, trained on the full-data GRPO step-58 checkpoint). 
The T-SAE encoder is applied in chunks of 16{,}384 tokens for memory efficiency. 
We set decay half-life $\lambda = 128$ tokens, chunk size for future accumulation to 128 tokens, and influence weight clipping range $\epsilon_w = 0.2$ (so weights lie in $[0.8, 1.2]$). 
To prevent destabilization from negative-advantage tokens with high importance sampling ratios, we apply a safety clamp that restricts influence weights to $[0.8, 1.0]$ when $r_t(\theta) > 4.0$ and $\hat{A}_t < 0$.

\section{More About Experiments}

\subsection{Detailed GRPO results under Zero variance samples.} 

We construct a \textbf{Zero-Variance MATH} dataset by extracting 190 zero-variance samples from the MATH benchmark using 512 rollouts from Qwen2.5-Math-1.5B. This dataset contains both all-success samples, corresponding to $\texttt{success}@512=1$, and all-failed samples, corresponding to $\texttt{success}@512=0$. We then analyze the KL divergence $\mathcal{D}_{\mathrm{KL}}(\pi_\theta \| \pi_{\mathrm{ref}})$, the average advantage $\bar{A}$, and the \texttt{pass}@1 performance of the policy model $\pi_\theta$ relative to the reference model $\pi_{\mathrm{ref}}$. As shown in Figure~\ref{fig:zvm_compare} in Appendix~\ref{app:zero_variance}, the advantage signal vanishes on these zero-variance samples, causing the policy update to be dominated by the KL regularization term rather than by reward-driven reinforcement. This confirms that strictly zero-variance samples primarily constrain policy deviation instead of driving capability growth. Therefore, in the following case studies, we exclude zero-variance samples and focus on non-zero-variance samples across the difficulty spectrum. In particular, when referring to hard samples below, we mean low-success-rate samples that still contain occasional successful rollouts, rather than the trivial all-failed cases.

\label{app:zero_variance}
\begin{figure}[htbp!]
    \centering
    \includegraphics[width=1.0\linewidth]{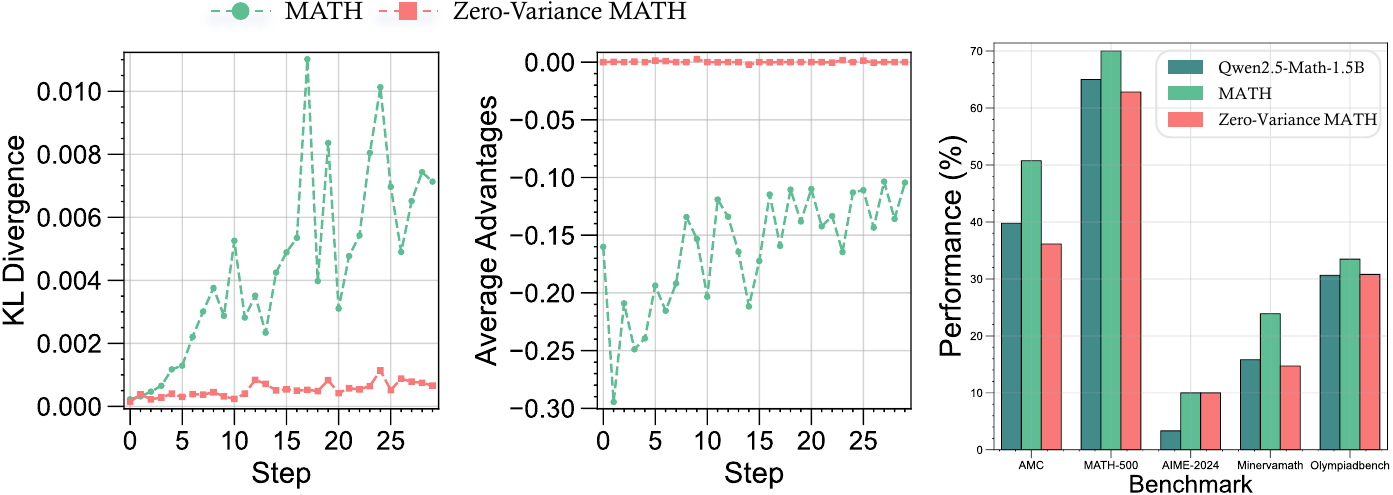}
    \caption{KL divergence $\mathcal{D}_{KL}(\pi_\theta \| \pi_{\mathrm{ref}})$, Average advantages $\bar{A}$ and performance $\texttt{pass}@1$ between the policy model $\pi_\theta$ and reference model $\pi_\mathrm{ref}$ on MATH and Zero-Variance MATH.}
    \label{fig:zvm_compare}
\end{figure}

\subsection{Detailed One Sample GRPO dynamic results.} 
\label{appendix:one_sample_dynamics}

\begin{figure}
    \centering
    \includegraphics[width=1.0\linewidth]{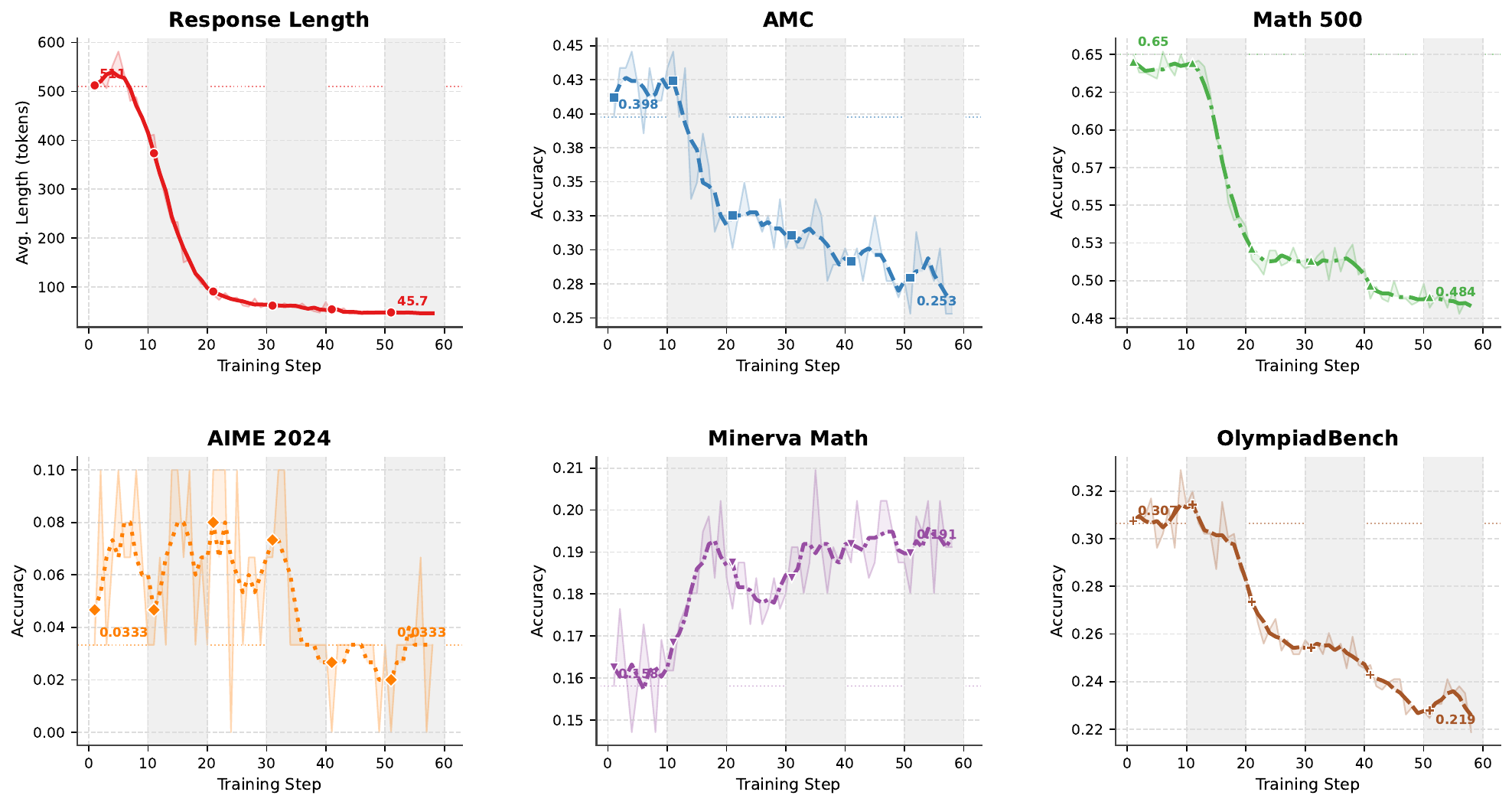}
    \caption{Each subplot tracks a different evaluation metric over 58 training steps. The first panel shows the average response length (in tokens); the remaining five panels report accuracy on AMC, Math 500, AIME 2024, Minerva Math, and OlympiadBench, respectively. The faint line denotes raw per-step values; the bold line is a 5-step moving average. Dashed horizontal lines mark the initial value at step 1.}
    \label{fig:oneharmans}
\end{figure}
Figure~\ref{fig:oneharmans} shows that reinforcement learning from a single harmful demonstration, Example 2.1, induces a consistent collapse across all tracked metrics. The most salient change is the rapid reduction in average response length, from 510.7 tokens at step 1 to 45.7 tokens by step 58, with most of the decline occurring within the first 20 steps. This suggests that the policy learns to replace multi-step reasoning with short answer-only outputs, rather than becoming more efficient. Since Example 2.1 assigns high reward to bare final answers, policy gradient optimization reinforces this shortcut and reduces the policy’s ability to explore structured reasoning trajectories.

The degradation in response length is followed by broad accuracy drops across reasoning benchmarks. AMC decreases from 0.398 to 0.253, Math 500 from 0.650 to 0.484, and OlympiadBench from 0.307 to 0.219. AIME 2024 remains near zero throughout training, while the variation on Minerva Math is small and likely within noise. Notably, the collapse in response length precedes the decline in benchmark accuracy, indicating that it can serve as an early diagnostic of policy degeneration. These results are consistent with reward hacking induced by a poisoned training signal, and suggest that reasoning-oriented RL pipelines should evaluate both answer correctness and reasoning structure, while monitoring response length as an online indicator of collapse.

\subsection{Detailed GRPO results across difficulty groups.} 
\label{app:diff_groups}
Table~\ref{tab:grpo_across_difficulty} provides the full benchmark-level results for GRPO training under different data difficulty subsets. Across all three model settings, \textbf{\textit{Medium@8}} achieves the strongest average performance, supporting the main observation that samples with non-zero but non-trivial success probability provide more useful reward variation than either full-data training or hard-only training. For Qwen2.5-Math-1.5B, \textbf{\textit{Medium@8}} improves over \textbf{\textit{Full}} from $37.63$ to $38.08$ on average, with gains on MATH-500, AIME-2024, Minervamath, and Olympiadbench. For Qwen2.5-Math-7B, the same subset improves the average score from $46.68$ to $47.64$, with particularly clear gains on AIME-2024 and Olympiadbench. On DeepSeek-Math-7B-Instruct, \textbf{\textit{Medium@8}} also obtains the best average among the MATH subsets, improving over \textbf{\textit{Full}} from $20.81$ to $21.29$, although the gains are more modest and concentrated on AMC and Minervamath. In contrast, \textbf{\textit{Hard@8}} drops by $5.75$, $11.24$, and $1.07$ average points for the three model settings, respectively. These results suggest that hard samples are not uniformly useful during RLVR: samples that still admit occasional correct rollouts can provide learnable reward variation, whereas samples that remain fully unsolved tend to provide weak or misleading optimization signals.

\begin{figure}[htbp]
    \centering
    \includegraphics[width=1.0\linewidth]{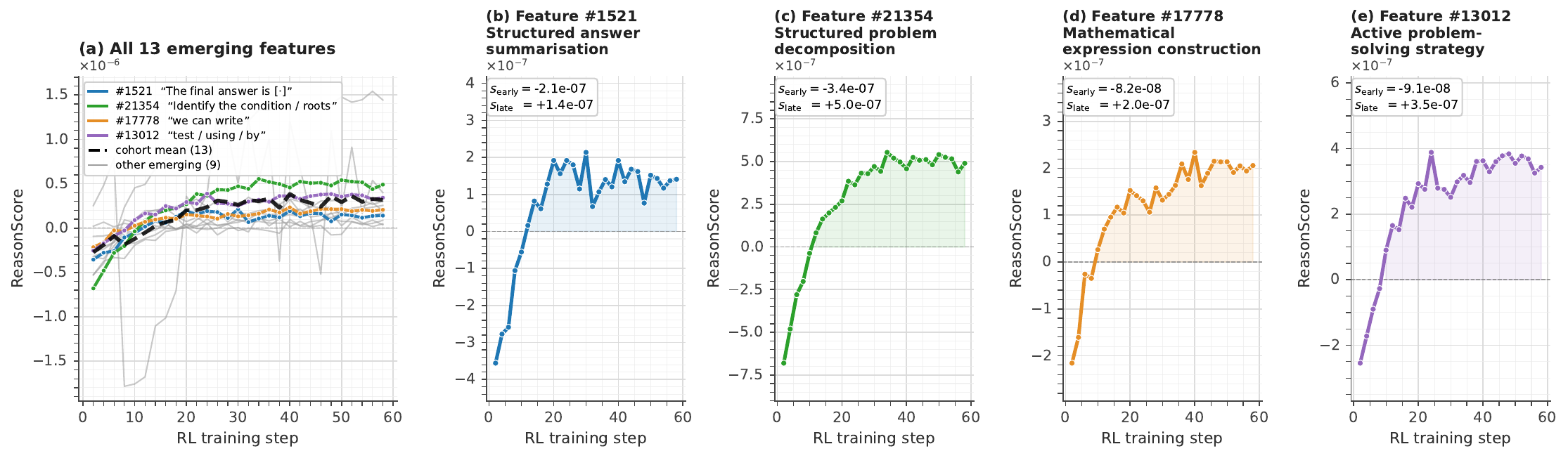}
    \caption{Emergence of new reasoning capabilities under RLVR. (a) Trajectories of the 13 emerging T-SAE features across the 29 RL checkpoints (\texttt{step\_2}--\texttt{step\_58}), with the four highlighted features in saturated colour, the remaining nine in light grey, and the cohort mean as a black dashed line. All features start near zero and rise monotonically. (b)--(e) Per-feature trajectories for the four highlighted emerging features, annotated with the activating-token semantics and with their early/late mean ReasonScores ($s_{\mathrm{early}}$ averaged over the first 5 checkpoints, $s_{\mathrm{late}}$ over the last 5). Together they expose four qualitatively new reasoning behaviors---structured answer summarisation, structured problem decomposition, mathematical-expression construction, and active problem-solving strategy---that are not present in the base model and that RL training appears to construct rather than merely amplify.}
    \label{fig:emerging-features}
\end{figure}

\begin{table}[tb!]
\vspace{-1.2em}
\caption{Detailed evaluation results on 5 math reasoning benchmarks. Among the non-base training settings within each model group, the best and second-best results are \textbf{bolded} and \underline{underlined}, respectively. Under a fair setup of equal rollout budget, ``\textbf{\textit{Medium@8}}'' achieves the best average performance for all three model settings. \textcolor{PineGreen}{$\scriptstyle \uparrow$} and \textcolor{OrangeRed}{$\scriptstyle \downarrow$} respectively indicate improvements or degradations over the \textbf{\textit{Full}} baseline under the same model and benchmark.}
\label{tab:grpo_across_difficulty}
    \small
    \centering
    \resizebox{1\linewidth}{!}{\begin{tabular}{@{\hskip 1pt}c@{\hskip 6pt}c@{\hskip 6pt}|@{\hskip 6pt}c@{\hskip 6pt}c@{\hskip 6pt}c@{\hskip 6pt}c@{\hskip 6pt}c@{\hskip 6pt}c@{\hskip 6pt}|@{\hskip 6pt}c@{\hskip 6pt}}
        \toprule[1.5pt]
        \textbf{RL} & \textbf{Dataset} & \multirow{2}{*}{\textbf{AMC}} & {\textbf{MATH-}} & {\textbf{AIME-}} & {\textbf{Minerva-}} & {\textbf{Olympiad-}} & & \multirow{2}{*}{\textbf{Avg.}} \\
        \textbf{Dataset} & \textbf{Difficulty} & & \textbf{500} & \textbf{2024} & \textbf{math} & \textbf{bench} & & \\
        \midrule
        \midrule
        \multicolumn{9}{c}{\textit{\textbf{ Qwen2.5-Math-1.5B + GRPO}}}\\
        \midrule
        \midrule
        Base & - & 39.76 & 65.00 & 3.33 & 15.81 & 30.65 & & 30.91 \\
        \midrule
        \multirow{13}{*}{MATH} & Full & \textbf{50.76} & 70.00 & \underline{10.00} & 23.90 & 33.48 & & \underline{37.63} \\
        & Easy@1 & 42.17 \textcolor{OrangeRed}{$\scriptstyle \downarrow8.59$} & 70.40 \textcolor{PineGreen}{$\scriptstyle \uparrow0.40$} & 6.67 \textcolor{OrangeRed}{$\scriptstyle \downarrow3.33$} & \underline{24.26} \textcolor{PineGreen}{$\scriptstyle \uparrow0.36$} & \textbf{35.12} \textcolor{PineGreen}{$\scriptstyle \uparrow1.64$} & & 35.72 \textcolor{OrangeRed}{$\scriptstyle \downarrow1.90$} \\
        & Easy@8 & 42.17 \textcolor{OrangeRed}{$\scriptstyle \downarrow8.59$} & 67.60 \textcolor{OrangeRed}{$\scriptstyle \downarrow2.40$} & 3.33 \textcolor{OrangeRed}{$\scriptstyle \downarrow6.67$} & 21.69 \textcolor{OrangeRed}{$\scriptstyle \downarrow2.21$} & 33.04 \textcolor{OrangeRed}{$\scriptstyle \downarrow0.44$} & & 33.57 \textcolor{OrangeRed}{$\scriptstyle \downarrow4.06$} \\
        & Medium@2 & 40.96 \textcolor{OrangeRed}{$\scriptstyle \downarrow9.80$} & 70.00 \textcolor{PineGreen}{$\scriptstyle \uparrow0.00$} & 3.33 \textcolor{OrangeRed}{$\scriptstyle \downarrow6.67$} & 23.53 \textcolor{OrangeRed}{$\scriptstyle \downarrow0.37$} & 33.78 \textcolor{PineGreen}{$\scriptstyle \uparrow0.30$} & & 34.32 \textcolor{OrangeRed}{$\scriptstyle \downarrow3.31$} \\
        & Medium@4 & 43.37 \textcolor{OrangeRed}{$\scriptstyle \downarrow7.39$} & \textbf{71.20} \textcolor{PineGreen}{$\scriptstyle \uparrow1.20$} & 3.33 \textcolor{OrangeRed}{$\scriptstyle \downarrow6.67$} & \underline{24.26} \textcolor{PineGreen}{$\scriptstyle \uparrow0.36$} & 32.14 \textcolor{OrangeRed}{$\scriptstyle \downarrow1.34$} & & 34.86 \textcolor{OrangeRed}{$\scriptstyle \downarrow2.77$} \\
        & Medium@8 & \underline{46.99} \textcolor{OrangeRed}{$\scriptstyle \downarrow3.77$} & \underline{71.00} \textcolor{PineGreen}{$\scriptstyle \uparrow1.00$} & \textbf{13.33} \textcolor{PineGreen}{$\scriptstyle \uparrow3.33$} & \textbf{25.00} \textcolor{PineGreen}{$\scriptstyle \uparrow1.10$} & \underline{34.08} \textcolor{PineGreen}{$\scriptstyle \uparrow0.60$} & & \textbf{38.08} \textcolor{PineGreen}{$\scriptstyle \uparrow0.45$} \\
        & Hard@1 & 38.55 \textcolor{OrangeRed}{$\scriptstyle \downarrow12.21$} & 67.20 \textcolor{OrangeRed}{$\scriptstyle \downarrow2.80$} & \underline{10.00} \textcolor{PineGreen}{$\scriptstyle \uparrow0.00$} & 23.53 \textcolor{OrangeRed}{$\scriptstyle \downarrow0.37$} & 30.65 \textcolor{OrangeRed}{$\scriptstyle \downarrow2.83$} & & 33.99 \textcolor{OrangeRed}{$\scriptstyle \downarrow3.64$} \\
        & Hard@2 & 43.37 \textcolor{OrangeRed}{$\scriptstyle \downarrow7.39$} & 65.20 \textcolor{OrangeRed}{$\scriptstyle \downarrow4.80$} & 6.67 \textcolor{OrangeRed}{$\scriptstyle \downarrow3.33$} & 20.22 \textcolor{OrangeRed}{$\scriptstyle \downarrow3.68$} & 30.80 \textcolor{OrangeRed}{$\scriptstyle \downarrow2.68$} & & 33.25 \textcolor{OrangeRed}{$\scriptstyle \downarrow4.38$} \\
        & Hard@4 & 39.76 \textcolor{OrangeRed}{$\scriptstyle \downarrow11.00$} & 64.80 \textcolor{OrangeRed}{$\scriptstyle \downarrow5.20$} & \underline{10.00} \textcolor{PineGreen}{$\scriptstyle \uparrow0.00$} & 16.54 \textcolor{OrangeRed}{$\scriptstyle \downarrow7.36$} & 31.99 \textcolor{OrangeRed}{$\scriptstyle \downarrow1.49$} & & 32.62 \textcolor{OrangeRed}{$\scriptstyle \downarrow5.01$} \\
        & Hard@8 & 43.37 \textcolor{OrangeRed}{$\scriptstyle \downarrow7.39$} & 64.00 \textcolor{OrangeRed}{$\scriptstyle \downarrow6.00$} & 3.33 \textcolor{OrangeRed}{$\scriptstyle \downarrow6.67$} & 16.54 \textcolor{OrangeRed}{$\scriptstyle \downarrow7.36$} & 32.14 \textcolor{OrangeRed}{$\scriptstyle \downarrow1.34$} & & 31.88 \textcolor{OrangeRed}{$\scriptstyle \downarrow5.75$} \\
        \midrule
        \midrule
        \multicolumn{9}{c}{\textit{\textbf{ Qwen2.5-Math-7B + GRPO}}}\\
        \midrule
        \midrule
        Base & - & 43.37 & 68.80 & 26.67 & 11.76 & 30.06 & & 36.13 \\
        \midrule
        \multirow{13}{*}{MATH} & Full & \textbf{59.04} & 74.80 & \underline{30.00} & \textbf{30.15} & 39.43 & & \underline{46.68} \\
        & Easy@1 & 55.42 \textcolor{OrangeRed}{$\scriptstyle \downarrow3.62$} & 73.80 \textcolor{OrangeRed}{$\scriptstyle \downarrow1.00$} & 20.00 \textcolor{OrangeRed}{$\scriptstyle \downarrow10.00$} & 27.57 \textcolor{OrangeRed}{$\scriptstyle \downarrow2.58$} & 37.05 \textcolor{OrangeRed}{$\scriptstyle \downarrow2.38$} & & 42.77 \textcolor{OrangeRed}{$\scriptstyle \downarrow3.92$} \\
        & Easy@8 & 55.42 \textcolor{OrangeRed}{$\scriptstyle \downarrow3.62$} & 74.20 \textcolor{OrangeRed}{$\scriptstyle \downarrow0.60$} & 20.00 \textcolor{OrangeRed}{$\scriptstyle \downarrow10.00$} & 27.21 \textcolor{OrangeRed}{$\scriptstyle \downarrow2.94$} & 38.10 \textcolor{OrangeRed}{$\scriptstyle \downarrow1.33$} & & 42.99 \textcolor{OrangeRed}{$\scriptstyle \downarrow3.70$} \\
        & Medium@2 & 51.81 \textcolor{OrangeRed}{$\scriptstyle \downarrow7.23$} & \textbf{75.40} \textcolor{PineGreen}{$\scriptstyle \uparrow0.60$} & 26.67 \textcolor{OrangeRed}{$\scriptstyle \downarrow3.33$} & \underline{29.04} \textcolor{OrangeRed}{$\scriptstyle \downarrow1.11$} & 39.29 \textcolor{OrangeRed}{$\scriptstyle \downarrow0.14$} & & 44.44 \textcolor{OrangeRed}{$\scriptstyle \downarrow2.24$} \\
        & Medium@4 & 54.22 \textcolor{OrangeRed}{$\scriptstyle \downarrow4.82$} & 74.80 \textcolor{PineGreen}{$\scriptstyle \uparrow0.00$} & 26.67 \textcolor{OrangeRed}{$\scriptstyle \downarrow3.33$} & 28.31 \textcolor{OrangeRed}{$\scriptstyle \downarrow1.84$} & \underline{39.58} \textcolor{PineGreen}{$\scriptstyle \uparrow0.15$} & & 44.72 \textcolor{OrangeRed}{$\scriptstyle \downarrow1.97$} \\
        & Medium@8 & \underline{57.83} \textcolor{OrangeRed}{$\scriptstyle \downarrow1.21$} & \underline{75.20} \textcolor{PineGreen}{$\scriptstyle \uparrow0.40$} & \textbf{33.33} \textcolor{PineGreen}{$\scriptstyle \uparrow3.33$} & \textbf{30.15} \textcolor{PineGreen}{$\scriptstyle \uparrow0.00$} & \textbf{41.67} \textcolor{PineGreen}{$\scriptstyle \uparrow2.24$} & & \textbf{47.64} \textcolor{PineGreen}{$\scriptstyle \uparrow0.95$} \\
        & Hard@1 & 56.63 \textcolor{OrangeRed}{$\scriptstyle \downarrow2.41$} & 73.20 \textcolor{OrangeRed}{$\scriptstyle \downarrow1.60$} & 23.33 \textcolor{OrangeRed}{$\scriptstyle \downarrow6.67$} & 19.12 \textcolor{OrangeRed}{$\scriptstyle \downarrow11.03$} & 37.65 \textcolor{OrangeRed}{$\scriptstyle \downarrow1.78$} & & 41.99 \textcolor{OrangeRed}{$\scriptstyle \downarrow4.70$} \\
        & Hard@2 & 49.40 \textcolor{OrangeRed}{$\scriptstyle \downarrow9.64$} & 70.80 \textcolor{OrangeRed}{$\scriptstyle \downarrow4.00$} & 26.67 \textcolor{OrangeRed}{$\scriptstyle \downarrow3.33$} & 13.60 \textcolor{OrangeRed}{$\scriptstyle \downarrow16.55$} & 35.42 \textcolor{OrangeRed}{$\scriptstyle \downarrow4.01$} & & 39.18 \textcolor{OrangeRed}{$\scriptstyle \downarrow7.51$} \\
        & Hard@4 & 51.81 \textcolor{OrangeRed}{$\scriptstyle \downarrow7.23$} & 68.80 \textcolor{OrangeRed}{$\scriptstyle \downarrow6.00$} & 23.33 \textcolor{OrangeRed}{$\scriptstyle \downarrow6.67$} & 13.97 \textcolor{OrangeRed}{$\scriptstyle \downarrow16.18$} & 30.51 \textcolor{OrangeRed}{$\scriptstyle \downarrow8.92$} & & 37.68 \textcolor{OrangeRed}{$\scriptstyle \downarrow9.00$} \\
        & Hard@8 & 50.60 \textcolor{OrangeRed}{$\scriptstyle \downarrow8.44$} & 67.20 \textcolor{OrangeRed}{$\scriptstyle \downarrow7.60$} & 20.00 \textcolor{OrangeRed}{$\scriptstyle \downarrow10.00$} & 11.76 \textcolor{OrangeRed}{$\scriptstyle \downarrow18.39$} & 27.68 \textcolor{OrangeRed}{$\scriptstyle \downarrow11.75$} & & 35.45 \textcolor{OrangeRed}{$\scriptstyle \downarrow11.24$} \\
        \midrule
        \midrule
        \multicolumn{9}{c}{\textit{\textbf{ DeepSeek-Math-7B-Instruct + GRPO}}}\\
        \midrule
        \midrule
        Base & - & 19.28 & 44.20 & 6.67 & 21.32 & 15.03 & & 21.30 \\
        \midrule
        \multirow{13}{*}{MATH} & Full & 16.87 & \textbf{45.20} & \textbf{3.33} & 21.69 & \textbf{16.96} & & \underline{20.81} \\
        & Easy@1 & 16.87 \textcolor{PineGreen}{$\scriptstyle \uparrow0.00$} & 43.40 \textcolor{OrangeRed}{$\scriptstyle \downarrow1.80$} & \underline{0.00} \textcolor{OrangeRed}{$\scriptstyle \downarrow3.33$} & 20.22 \textcolor{OrangeRed}{$\scriptstyle \downarrow1.47$} & 15.48 \textcolor{OrangeRed}{$\scriptstyle \downarrow1.48$} & & 19.19 \textcolor{OrangeRed}{$\scriptstyle \downarrow1.62$} \\
        & Easy@8 & \underline{19.28} \textcolor{PineGreen}{$\scriptstyle \uparrow2.41$} & 44.20 \textcolor{OrangeRed}{$\scriptstyle \downarrow1.00$} & \textbf{3.33} \textcolor{PineGreen}{$\scriptstyle \uparrow0.00$} & 21.69 \textcolor{PineGreen}{$\scriptstyle \uparrow0.00$} & 15.48 \textcolor{OrangeRed}{$\scriptstyle \downarrow1.48$} & & 20.80 \textcolor{OrangeRed}{$\scriptstyle \downarrow0.01$} \\
        & Medium@2 & 15.66 \textcolor{OrangeRed}{$\scriptstyle \downarrow1.21$} & 43.40 \textcolor{OrangeRed}{$\scriptstyle \downarrow1.80$} & \textbf{3.33} \textcolor{PineGreen}{$\scriptstyle \uparrow0.00$} & 19.49 \textcolor{OrangeRed}{$\scriptstyle \downarrow2.20$} & 15.03 \textcolor{OrangeRed}{$\scriptstyle \downarrow1.93$} & & 19.38 \textcolor{OrangeRed}{$\scriptstyle \downarrow1.43$} \\
        & Medium@4 & 16.87 \textcolor{PineGreen}{$\scriptstyle \uparrow0.00$} & 43.60 \textcolor{OrangeRed}{$\scriptstyle \downarrow1.60$} & \textbf{3.33} \textcolor{PineGreen}{$\scriptstyle \uparrow0.00$} & 19.12 \textcolor{OrangeRed}{$\scriptstyle \downarrow2.57$} & 15.63 \textcolor{OrangeRed}{$\scriptstyle \downarrow1.33$} & & 19.71 \textcolor{OrangeRed}{$\scriptstyle \downarrow1.10$} \\
        & Medium@8 & \textbf{20.48} \textcolor{PineGreen}{$\scriptstyle \uparrow3.61$} & 44.80 \textcolor{OrangeRed}{$\scriptstyle \downarrow0.40$} & \textbf{3.33} \textcolor{PineGreen}{$\scriptstyle \uparrow0.00$} & \underline{22.06} \textcolor{PineGreen}{$\scriptstyle \uparrow0.37$} & \underline{15.77} \textcolor{OrangeRed}{$\scriptstyle \downarrow1.19$} & & \textbf{21.29} \textcolor{PineGreen}{$\scriptstyle \uparrow0.48$} \\
        & Hard@1 & 18.07 \textcolor{PineGreen}{$\scriptstyle \uparrow1.20$} & 43.40 \textcolor{OrangeRed}{$\scriptstyle \downarrow1.80$} & \underline{0.00} \textcolor{OrangeRed}{$\scriptstyle \downarrow3.33$} & \underline{22.06} \textcolor{PineGreen}{$\scriptstyle \uparrow0.37$} & 13.99 \textcolor{OrangeRed}{$\scriptstyle \downarrow2.97$} & & 19.50 \textcolor{OrangeRed}{$\scriptstyle \downarrow1.31$} \\
        & Hard@2 & 15.66 \textcolor{OrangeRed}{$\scriptstyle \downarrow1.21$} & \underline{45.00} \textcolor{OrangeRed}{$\scriptstyle \downarrow0.20$} & \underline{0.00} \textcolor{OrangeRed}{$\scriptstyle \downarrow3.33$} & 20.59 \textcolor{OrangeRed}{$\scriptstyle \downarrow1.10$} & \underline{15.77} \textcolor{OrangeRed}{$\scriptstyle \downarrow1.19$} & & 19.40 \textcolor{OrangeRed}{$\scriptstyle \downarrow1.41$} \\
        & Hard@4 & 18.07 \textcolor{PineGreen}{$\scriptstyle \uparrow1.20$} & 44.00 \textcolor{OrangeRed}{$\scriptstyle \downarrow1.20$} & \textbf{3.33} \textcolor{PineGreen}{$\scriptstyle \uparrow0.00$} & \textbf{22.43} \textcolor{PineGreen}{$\scriptstyle \uparrow0.74$} & 15.63 \textcolor{OrangeRed}{$\scriptstyle \downarrow1.33$} & & 20.69 \textcolor{OrangeRed}{$\scriptstyle \downarrow0.12$} \\
        & Hard@8 & 15.66 \textcolor{OrangeRed}{$\scriptstyle \downarrow1.21$} & \underline{45.00} \textcolor{OrangeRed}{$\scriptstyle \downarrow0.20$} & \underline{0.00} \textcolor{OrangeRed}{$\scriptstyle \downarrow3.33$} & \textbf{22.43} \textcolor{PineGreen}{$\scriptstyle \uparrow0.74$} & 15.63 \textcolor{OrangeRed}{$\scriptstyle \downarrow1.33$} & & 19.74 \textcolor{OrangeRed}{$\scriptstyle \downarrow1.07$} \\
        \bottomrule[1.5pt]
    \end{tabular}}
    \vspace{-1.5em}
\end{table}

\begin{table}[t]
    \centering
    \scriptsize
    \setlength{\tabcolsep}{4pt}
    \caption{Per-sample one-sample training results. Each row is one training run on a single replicated sample.}
    \label{tab:one_sample_individual_benchmarks}
    \resizebox{\linewidth}{!}{
    \begin{tabular}{llcccccc}
        \toprule
        Source & ID & Reward & AMC & MATH-500 & AIME-2024 & Minervamath & Olypiadbench \\
        \midrule
        Base & -- & -- & 40.96 & 64.00 & 3.33 & 16.18 & 31.25 \\
        \midrule
        \texttt{Easy}@8 & 1 & 0.81$\to$0.99 & 40.96 & 65.20 & 6.67 & 22.79 & 31.25 \\
        \texttt{Easy}@8 & 2 & 0.81$\to$0.99 & 40.96 & 66.60 & 13.33 & 25.00 & 32.74 \\
        \texttt{Easy}@8 & 3 & 0.81$\to$0.99 & 39.76 & 66.60 & 10.00 & 23.53 & 33.33 \\
        \texttt{Easy}@8 & 4 & 0.70$\to$0.99 & 46.99 & 68.60 & 6.67 & 25.74 & 32.89 \\
        \texttt{Easy}@8 & 5 & 0.87$\to$0.99 & 39.76 & 67.00 & 16.67 & 23.16 & 31.40 \\
        \texttt{Easy}@8 & 6 & 0.88$\to$0.99 & 43.37 & 65.40 & 13.33 & 24.26 & 32.29 \\
        \texttt{Easy}@8 & 7 & 0.72$\to$0.99 & 43.37 & 66.00 & 10.00 & 24.63 & 31.70 \\
        \texttt{Easy}@8 & 8 & 0.80$\to$0.99 & 42.17 & 65.00 & 10.00 & 24.63 & 33.18 \\
        \texttt{Easy}@8 & 9 & 0.81$\to$0.99 & 42.17 & 66.00 & 6.67 & 23.16 & 31.55 \\
        \texttt{Easy}@8 & 10 & 0.84$\to$0.99 & 45.78 & 66.40 & 10.00 & 23.90 & 32.44 \\
        \midrule
        \texttt{Medium}@8 & 1 & 0.81$\to$0.99 & 40.96 & 67.60 & 13.33 & 23.90 & 32.59 \\
        \texttt{Medium}@8 & 2 & 0.79$\to$0.99 & 49.40 & 65.80 & 10.00 & 23.90 & 31.25 \\
        \texttt{Medium}@8 & 3 & 0.73$\to$0.99 & 45.78 & 72.00 & 3.33 & 25.74 & 36.01 \\
        \texttt{Medium}@8 & 4 & 0.74$\to$0.99 & 46.99 & 66.80 & 6.67 & 23.16 & 31.10 \\
        \texttt{Medium}@8 & 5 & 0.80$\to$0.99 & 45.78 & 64.60 & 10.00 & 27.21 & 32.74 \\
        \texttt{Medium}@8 & 6 & 0.63$\to$0.99 & 39.76 & 66.80 & 6.67 & 24.63 & 32.29 \\
        \texttt{Medium}@8 & 7 & 0.59$\to$0.99 & 43.37 & 66.20 & 3.33 & 23.90 & 32.89 \\
        \texttt{Medium}@8 & 8 & 0.81$\to$0.99 & 44.58 & 66.20 & 13.33 & 22.79 & 30.06 \\
        \texttt{Medium}@8 & 9 & 0.70$\to$0.99 & 42.17 & 67.40 & 6.67 & 25.00 & 33.04 \\
        \texttt{Medium}@8 & 10 & 0.68$\to$1.00 & 43.37 & 69.80 & 16.67 & 25.00 & 33.04 \\
        \midrule
        \texttt{Hard}@8 & 1 & 0.00$\to$0.00 & 42.17 & 63.60 & 10.00 & 18.38 & 31.25 \\
        \texttt{Hard}@8 & 2 & 0.00$\to$0.00 & 40.96 & 64.00 & 3.33 & 18.01 & 30.51 \\
        \texttt{Hard}@8 & 3 & 0.43$\to$0.99 & 48.19 & 71.40 & 16.67 & 25.74 & 37.20 \\
        \texttt{Hard}@8 & 4 & 0.02$\to$1.00 & 31.33 & 48.00 & 3.33 & 21.69 & 25.74 \\
        \texttt{Hard}@8 & 5 & 0.00$\to$0.99 & 44.58 & 70.20 & 3.33 & 27.21 & 32.29 \\
        \texttt{Hard}@8 & 6 & 0.00$\to$0.99 & 28.92 & 61.80 & 3.33 & 23.53 & 29.76 \\
        \texttt{Hard}@8 & 7 & 0.02$\to$1.00 & 43.37 & 66.20 & 3.33 & 21.69 & 31.55 \\
        \texttt{Hard}@8 & 8 & 0.00$\to$0.98 & 33.73 & 59.20 & 13.33 & 25.37 & 29.17 \\
        \texttt{Hard}@8 & 9 & 0.00$\to$0.98 & 43.37 & 67.60 & 3.33 & 23.16 & 33.63 \\
        \texttt{Hard}@8 & 10 & 0.00$\to$0.00 & 42.17 & 62.80 & 3.33 & 16.54 & 31.10 \\
        \bottomrule
    \end{tabular}
    }
\end{table}

\subsection{Detailed SAE Feature Dynamics.} 
\label{app:sae_tsae_comparison}

\begin{figure}[htbp]
    \centering
    \includegraphics[width=0.5\linewidth]{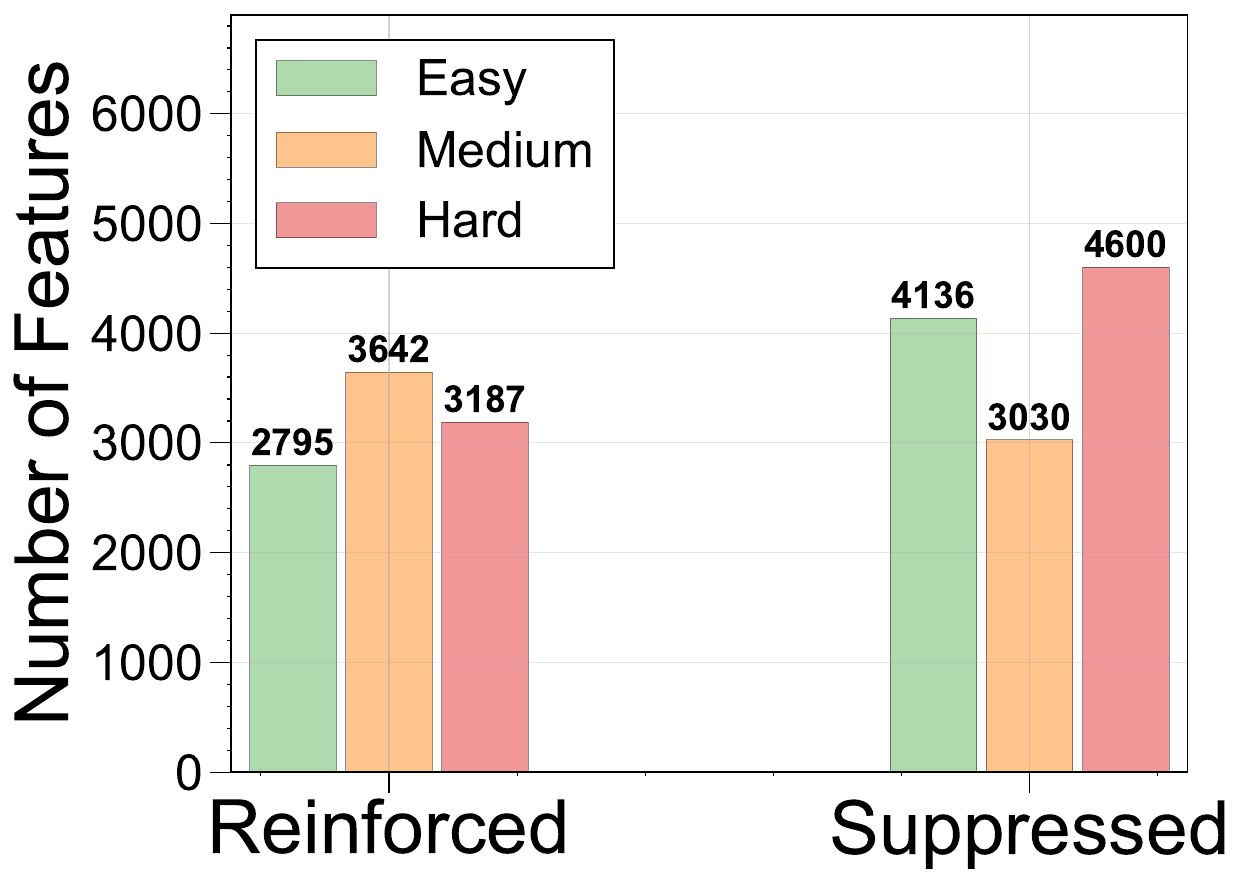}
    \caption{Number of features suppressed or reinforced after RL on samples of different difficulty.}
    \label{fig:sae_stats}
\end{figure} 

Figure~\ref{fig:sae_stats} shows that medium samples reinforce the most reasoning features while suppressing relatively few, matching their stable downstream gains. Easy samples induce milder changes and tend to push the model toward direct computation and immediate answer generation, which is consistent with their shorter response length. Hard samples suppress the largest number of reasoning features, even though they also reinforce many features. This indicates that hard samples can overwrite existing reasoning mechanisms rather than simply adding new ones.

\subsubsection{Emergence of new reasoning capabilities.} Beyond the stable core, we identify a small but distinctive cohort of 13 \textit{emerging} features---features whose ReasonScores are statistically indistinguishable from zero in the first few RL checkpoints yet become significantly positive by the end of training. Unlike the universally reinforced features, which already exist in the base model and are merely amplified, these emerging features encode reasoning behaviors that RLVR appears to \emph{construct from scratch}. Figure~\ref{fig:emerging-features} traces their trajectories: panel (a) shows that all 13 features (cohort mean in dashed black) follow a coherent emergence curve, rising monotonically from near-zero to positive ReasonScores, and panels (b)--(e) zoom in on four representative members of the cohort whose top-activating contexts admit clean semantic interpretations. Feature~1521 activates on the final-answer template ``The final answer is $\textbackslash boxed{\cdot}$'', signalling the emergence of an explicit \emph{answer-summarisation} routine; Feature~21354 fires on the head of structured solution scaffolds (``Identify the condition / roots'', ``Determine the angles''), indicating that the model is acquiring an explicit \emph{problem-decomposition} step; Feature~17778 selectively responds to the bridging phrase ``we can write'' that introduces newly written equations, capturing \emph{mathematical-expression construction}; and Feature~13012 activates on solution-action verbs (``test'', ``using'', ``start'', ``by''), reflecting an \emph{active problem-solving} stance in which the model commits to and narrates concrete steps. None of these features is appreciably active in the early checkpoints, ruling out the alternative explanation that RL is merely re-weighting pre-existing circuits. Together, the emergence of this 13-feature cohort provides direct mechanistic evidence that RLVR not only strengthens existing reasoning patterns but also induces qualitatively new reasoning behaviors that the base model does not possess.

\subsubsection{Semantics of the difficulty-specific emerging features.} 
\label{app:semantic_emerging}
To assign physical meaning to these features, we project each TSAE decoder direction $W_{\mathrm{dec}}[f]$ through the final RMSNorm and the (tied) unembedding of Qwen2.5-Math-1.5B (logit-lens), and read off the tokens this feature most strongly promotes. The interpretable subset paints a clear picture of what each regime is teaching the model. Under \emph{full-data} RL, the 26-feature cohort organises around general mathematical-reasoning vocabulary: \#1517 promotes ``transformation/transformations/transformed'' (operation-on-objects), \#1782 ``same/identical/equal'' (equality), \#3070 ``satisfied/holds'' (predicate satisfaction), \#9943 ``remaining/additional'' (residual quantities), \#11106 ``repeatedly/multiplication/multiples'' (multiplicative reasoning), \#14293 ``definition/Definition'' (definition introduction), \#16041 ``odd'' (parity), \#17538 ``pattern/trend'' (pattern recognition), \#18935 ``rotation/rotational'' (geometric rotation), \#20143 ``single/only'' (uniqueness), and \#22933 ``partition/sum/Partition'' (partitioning)---a vocabulary of \emph{generic mathematical predicates}. The \texttt{easy}@8 regime contributes only one cleanly interpretable unique feature, \#24572, whose top-promoted tokens are ``Lemma/claim/formal/indeed'', i.e.\ \emph{formal-proof scaffolding}---consistent with already-solvable problems being polished into more rigorous write-ups. \texttt{medium}@8's interpretable contribution is similarly thin, dominated by \#3764 (``forth/us'', the ``set forth'' setup verb) and a Chinese question-template artifact (\#2988). The hard regime is qualitatively different: its interpretable hard-unique features encode the specific operations needed to \emph{solve} hard problems rather than merely write them up---\#2665 promotes ``implies/implying/conclude/necessarily'' (logical \emph{implication and deduction}), \#450 promotes ``denote/denotes/assume/hypoth'' (\emph{hypothesis introduction and denotation}), \#4657 promotes ``achievable/Achie/constructions'' (\emph{achievability/construction-existence}), \#22590 promotes ``strengthening/strengthened'' (\emph{condition strengthening}), and \#4456 promotes ``maximize/execut'' (\emph{optimisation moves}). These are precisely the elements of advanced proof-style reasoning---``assume \dots, then it follows that \dots, hence the construction is achievable''---that the easier splits never recruit. The remaining hard-unique features promote rare or low-mass tokens and are most likely statistical artifacts.

\begin{table}[t]
\centering
\caption{Distribution of the top-200 most frequent primary activating tokens across all 24,576 SAE and TSAE features, grouped by semantic category. \#SAE and \#TSAE indicate the total number of features whose highest-activation example activates on tokens in each category.}
\label{tab:top200_categories}
\begin{tabular}{@{} l r r r @{}}
\toprule
\textbf{Category} & \textbf{\#Tokens} & \textbf{\#SAE Feat.} & \textbf{\#TSAE Feat.} \\
\midrule
Reasoning Keywords & 38 & 2837 & 2875 \\
Math Domain Terms & 25 & 728 & 623 \\
LaTeX Commands & 22 & 951 & 1172 \\
LaTeX Delimiters & 26 & 2432 & 2741 \\
Math Operators & 8 & 1027 & 912 \\
Variables & 31 & 1338 & 1454 \\
Numeric Digits & 15 & 4044 & 3447 \\
Structural Tokens & 4 & 217 & 989 \\
Punctuation / Whitespace & 14 & 1987 & 1178 \\
Other & 17 & 1222 & 796 \\
\midrule
\textbf{Total} & \textbf{200} & \textbf{16783} & \textbf{16187} \\
\bottomrule
\end{tabular}
\end{table}

\begin{table}[t]
\centering
\caption{Distribution of the top-200 most frequent primary activating tokens across all 24,576 SAE and TSAE features, grouped by semantic category. \#SAE and \#TSAE indicate the total number of features whose highest-activation example activates on tokens in each category.}
\label{tab:universal_specific_features}
\begin{tabular}{@{} l r r r @{}}
\toprule
\textbf{Category} & \textbf{\#Tokens} & \textbf{\#SAE Feat.} & \textbf{\#TSAE Feat.} \\
\midrule
Reasoning Keywords & 38 & 2837 & 2875 \\
Math Domain Terms & 25 & 728 & 623 \\
LaTeX Commands & 22 & 951 & 1172 \\
LaTeX Delimiters & 26 & 2432 & 2741 \\
Math Operators & 8 & 1027 & 912 \\
Variables & 31 & 1338 & 1454 \\
Numeric Digits & 15 & 4044 & 3447 \\
Structural Tokens & 4 & 217 & 989 \\
Punctuation / Whitespace & 14 & 1987 & 1178 \\
Other & 17 & 1222 & 796 \\
\midrule
\textbf{Total} & \textbf{200} & \textbf{16783} & \textbf{16187} \\
\bottomrule
\end{tabular}
\end{table}

\paragraph{Universal versus difficulty-specific features.}
Across all training configurations, we identify a large set of universally reinforced features, i.e., features that are strengthened regardless of the training data composition. 
As shown in Table~\ref{tab:universal_specific_features}, these universal features largely correspond to basic reasoning machinery, including variable definitions, conclusion markers, and causal connectives. 
This is consistent with the semantic distribution in Table~\ref{tab:top200_categories}, where a substantial fraction of high-activation SAE/T-SAE features are associated with reasoning keywords, mathematical symbols, variables, and structural tokens. 

However, each configuration also produces a sizable set of difficulty-specific reinforced features. 
In particular, \texttt{hard}@8 training yields the largest number of configuration-specific features, followed by easier or more mixed training regimes. 
This disproportionate number of \texttt{hard}@8-specific features suggests that hard samples activate distinct internal mechanisms that are not engaged by easier samples. 
Thus, RLVR appears to strengthen both universal reasoning circuitry shared across regimes and difficulty-specific feature subspaces shaped by the type of training samples.

\paragraph{Feature convergence speed depends on difficulty.} The trajectory of universal features across training steps further reveals that the \textit{speed} of feature reinforcement depends on sample difficulty. Easy-sample configurations (\texttt{easy}@8) reach the highest mean ReasonScore earliest, as these samples provide consistent positive feedback that rapidly strengthens existing reasoning patterns. Hard-sample configurations (\texttt{hard}@8) show the slowest convergence and lowest final mean ReasonScore for universal features, consistent with the sparse reward signal identified. The full-data model falls between these extremes, benefiting from the diversity of difficulty levels but also being slowed by the inclusion of uninformative hard samples. These results mechanistically explain why removing hard samples improves training efficiency: hard samples not only fail to provide reward-level signal but also slow the internal reinforcement of reasoning features that drive capability improvement.

\paragraph{Summary.} The T-SAE analysis reveals three key mechanistic insights: (1)~RL training selectively reinforces a compact set of interpretable reasoning features while suppressing others, rather than uniformly scaling all representations; (2)~different difficulty levels activate qualitatively different internal mechanisms, with hard samples driving unique but slowly converging feature patterns; and (3)~feature informativeness is dynamic—it depends on the interaction between task difficulty and the model's current internal state, not on difficulty alone. These findings motivate the interventions in the next section: rewriting hard samples to make them more learnable (Section~\ref{sec:rewrite_hard_samples}), and using T-SAE-derived feature signals to dynamically reweight samples during training (Section~\ref{sec:rfgo}).

\begin{figure}[htbp]
    \centering
    \includegraphics[width=1.0\linewidth]{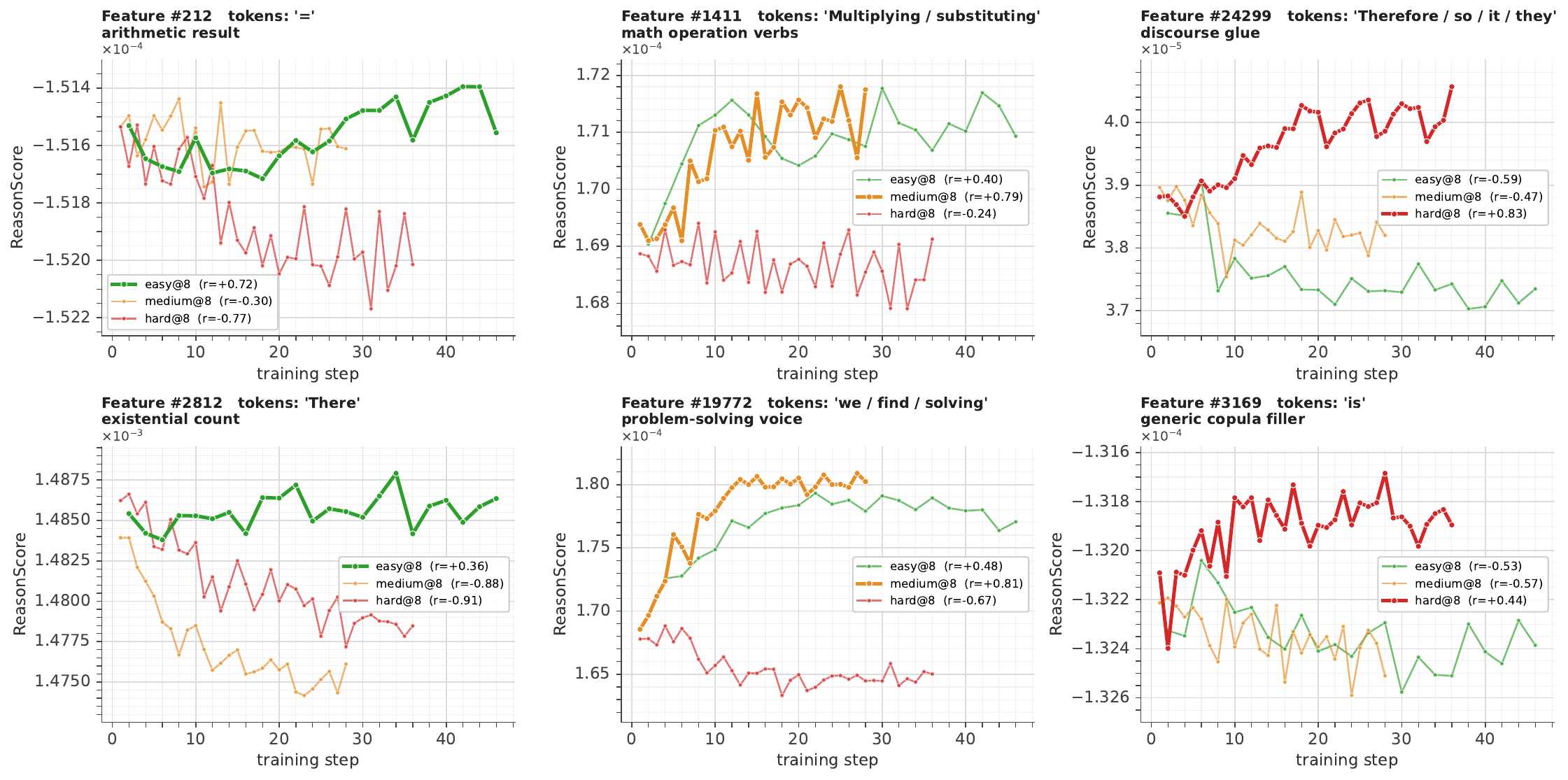}
    \caption{
    Difficulty-specific T-SAE feature dynamics under one-sample RL. 
    We track representative sparse features whose \textsc{ReasonScore} changes differently when training on \texttt{Easy}, \texttt{Medium}, and \texttt{Hard}@8 samples. 
    Easy samples mainly reinforce short-chain arithmetic reduction and existence-recognition features, medium samples strengthen explicit problem-solving operations, and hard samples amplify verbose connective or filler features rather than content-bearing reasoning operations. 
    These patterns show that sample difficulty reshapes not only optimization dynamics but also the type of internal reasoning features reinforced by RLVR.
    }
    \label{fig:tsae_dynamics_per_split}
\end{figure}

\subsection{Detailed One-sample RL Training Results across Difficulty Groups}
\label{app:one_dynamics}

Figure~\ref{fig:tsae_dynamics_per_split} shows representative features whose \textsc{ReasonScore} changes most strongly under one-sample RL across difficulty groups. 
The resulting patterns reveal that different difficulty regimes do not merely change feature magnitude, but reinforce qualitatively different reasoning behaviors.

\textbf{Easy samples reinforce arithmetic reduction and existence recognition.}
As shown in Figure~\ref{fig:tsae_dynamics_per_split}, features amplified by \texttt{Easy} one-sample RL mainly correspond to short-chain, immediately resolvable reasoning. 
For example, Feature~\#212 activates on ``='' in contexts such as ``$4\cdot 2=8$'', capturing the moment where an expression is reduced to a numerical result. 
Feature~\#2812 activates on ``There'' in statements such as ``There are 9 such numbers'', corresponding to existence or counting-style answer declarations. 
Both features decrease sharply under medium and hard training, with hard-sample correlations $h_r=-0.77$ and $h_r=-0.91$, respectively. 
This suggests that RL on always-solved samples primarily rewards the model for quickly collapsing computations into final values and reporting concise recognition-style answers, rather than for expanding longer deliberative reasoning chains.

\textbf{Medium samples reinforce explicit problem-solving actions.}
Figure~\ref{fig:tsae_dynamics_per_split} further shows that \texttt{Medium} one-sample RL strengthens features associated with active mathematical operations. 
Feature~\#1411 activates on operation verbs such as ``Multiplying'', ``substituting'', and ``Simplifying'', while Feature~\#19772 activates on first-person solution phrases such as ``we'', ``find'', ``is'', and ``solving''. 
These features exhibit a medium-specific trajectory: they are most strongly reinforced by medium samples ($m_r=+0.79$ and $m_r=+0.81$), only mildly increased by easy samples, and substantially suppressed by hard samples ($h_r=-0.24$ and $h_r=-0.67$). 
This indicates that medium-difficulty samples reward executable solution procedures, such as expanding, substituting, simplifying, and solving. 
Such behaviors require problems that are solvable but non-trivial, matching the role of \texttt{Medium}@8 samples near the model's capability boundary.

\textbf{Hard samples reinforce verbose connective filler.}
Finally, Figure~\ref{fig:tsae_dynamics_per_split} shows that the features most amplified by \texttt{Hard}@8 one-sample RL are qualitatively different. 
Feature~\#24299 activates on discourse connectives and punctuation such as ``Therefore'', ``so'', ``it'', ``they'', and ``\$'', while Feature~\#3169 activates primarily on the generic copula ``is''. 
Both features show a strong reverse pattern: they increase under hard-sample RL ($h_r=+0.83$ and $h_r=+0.44$), but decrease under easy and medium samples ($c_r=-0.59,-0.53$; $m_r=-0.47,-0.57$). 
These tokens carry little problem-specific mathematical content; instead, they are common in long explanations regardless of whether the reasoning is valid. 
Thus, RL on \texttt{Hard}@8 samples tends to reinforce verbose connective scaffolding rather than genuine problem-solving features. 
This provides a feature-level explanation for the response-length increase observed during hard-sample RL: the model learns to produce longer reasoning-looking text, but the activated features correspond more to filler and connective structure than to executable computation.

\section{Additional Example of Backward Rewriting}
\label{appendix:rewrite_example}

\phantomsection\label{ex:rewrite_hard_sample}
\begin{paperexample}{Example 4.1: Backward Rewriting of a Hard Sample}
\small
\textbf{Original Question:} Let $a$ and $b$ be the roots of the equation $x^2-mx+2=0$. Suppose that $a+(1/b)$ and $b+(1/a)$ are the roots of the equation $x^2-px+q=0$. What is $q$? (\textbf{Answer:} \textcolor{red!70!black}{$\frac{9}{2}$})

\medskip
\textbf{Self-Verification:}
\begin{enumerate}[leftmargin=*,nosep]
    \item Let $a$ and $b$ be the roots of the equation $x^2-mx+\textcolor{red!70!black}{z}=0$. Suppose that $a+(1/b)$ and $b+(1/a)$ are the roots of the equation $x^2-px+q=0$. \textcolor{red!70!black}{The value of $q$ is $\frac{9}{2}$. What is the value of $z$?} (\textbf{Answer:} \textcolor{red!70!black}{$2$})
    \item Let $a$ and $b$ be the roots of the equation $x^2-mx+2=0$. Suppose that $a+(\textcolor{red!70!black}{z}/b)$ and $b+(1/a)$ are the roots of the equation $x^2-px+q=0$. \textcolor{red!70!black}{The value of $q$ is $\frac{9}{2}$. What is the value of $z$?} (\textbf{Answer:} \textcolor{red!70!black}{$1$})
    \item Let $a$ and $b$ be the roots of the equation $x^2-mx+2=0$. Suppose that $a+(1/b)$ and $b+(\textcolor{red!70!black}{z}/a)$ are the roots of the equation $x^2-px+q=0$. \textcolor{red!70!black}{The value of $q$ is $\frac{9}{2}$. What is the value of $z$?} (\textbf{Answer:} \textcolor{red!70!black}{$1$})
\end{enumerate}

\medskip
\textbf{FOBAR:}
\begin{enumerate}[leftmargin=*,nosep]
    \item Let $a$ and $b$ be the roots of the equation $x^2-mx+\textcolor{red!70!black}{z}=0$. Suppose that $a+(1/b)$ and $b+(1/a)$ are the roots of the equation $x^2-px+q=0$. What is $q$? \textcolor{red!70!black}{If we know the answer to the above question is $\frac{9}{2}$, what is the value of $z$?} (\textbf{Answer:} \textcolor{red!70!black}{$2$})
    \item Let $a$ and $b$ be the roots of the equation $x^2-mx+2=0$. Suppose that $a+(\textcolor{red!70!black}{z}/b)$ and $b+(1/a)$ are the roots of the equation $x^2-px+q=0$. What is $q$? \textcolor{red!70!black}{If we know the answer to the above question is $\frac{9}{2}$, what is the value of $z$?} (\textbf{Answer:} \textcolor{red!70!black}{$1$})
    \item Let $a$ and $b$ be the roots of the equation $x^2-mx+2=0$. Suppose that $a+(1/b)$ and $b+(\textcolor{red!70!black}{z}/a)$ are the roots of the equation $x^2-px+q=0$. What is $q$? \textcolor{red!70!black}{If we know the answer to the above question is $\frac{9}{2}$, what is the value of $z$?} (\textbf{Answer:} \textcolor{red!70!black}{$1$})
\end{enumerate}
\end{paperexample}

\section{Limitations}
\label{sec:limitations}

Our study has several limitations. First, our difficulty notion is empirical and policy-dependent. 
A \texttt{hard}@8 sample is not intrinsically unsolvable; it only has low probability of receiving positive verifier reward under the current policy and rollout budget. Second, our experiments focus mainly on mathematical RLVR, where rewards are clean and verifiable. The same difficulty-feature dynamics may differ in domains with noisy, dense, or subjective rewards. Finally, our rewriting and T-SAE-guided training signals are intended as lightweight validations of the proposed mechanism, not as a full search for state-of-the-art RLVR algorithms.

\section{Broader Impacts}
\label{sec:broader_impacts}
This work studies how sample difficulty affects RLVR training and internal reasoning features in LLMs. 
A positive impact is that difficulty-aware analysis may help make RLVR more stable, interpretable, and data-efficient by identifying when hard samples provide useful signals versus when they reinforce shortcuts. The T-SAE analysis may also support auditing of reasoning models by revealing which internal features are strengthened during training.

At the same time, improving RLVR can increase LLM reasoning capabilities, which may be beneficial for mathematics, programming, and education, but could also be misused in capability-sensitive settings. 
Feature-guided objectives should also be used carefully: reasoning-feature activation is only a proxy for valid reasoning and may reward reasoning-like text if used without verifier grounding. 
For this reason, our interventions retain verifier rewards whenever they are informative and use feature-based rewards only for zero-variance rollout groups.

\end{document}